\documentclass{article}

\usepackage{PRIMEarxiv}

\usepackage[utf8]{inputenc} 
\usepackage[T1]{fontenc}    
\usepackage{graphicx}%
\usepackage{multirow}%
\usepackage{amsmath,amssymb,amsfonts}%
\usepackage{amsthm}%
\usepackage{mathrsfs}%
\usepackage[title]{appendix}%
\usepackage{xcolor}%
\usepackage{textcomp}%
\usepackage{manyfoot}%
\usepackage{booktabs}%
\usepackage{algorithm}%
\usepackage{algorithmicx}%
\usepackage{algpseudocode}%
\usepackage{listings}%

\usepackage{xspace}
\usepackage{hyperref}
\usepackage{multirow}
\usepackage{tablefootnote}
\usepackage{caption}
\usepackage{subcaption}
\usepackage{array}
\usepackage[acronym]{glossaries}

\graphicspath{{media/}}     

\title{A Survey of the Impact of Self-Supervised Pretraining for Diagnostic Tasks with Radiological Images
}

\author{
  Blake VanBerlo\thanks{B. VanBerlo is a Vanier Scholar (FRN 186945) supported by the Natural Sciences and Engineering Research Council of Canada (NSERC).}  \\
  Cheriton School of Computer Science \\
  University of Waterloo \\
  Waterloo, Canada\\
  \texttt{bvanberl@uwaterloo.ca} \\
   \And
  Jesse Hoey \\
  Cheriton School of Computer Science \\
  University of Waterloo \\
  Waterloo, Canada\\
  \And
  Alexander Wong \\
  Department of Systems Design Engineering \\
  University of Waterloo \\
  Waterloo, Canada\\
}
\newcommand{\etal}{\textit{et~al.}}
\newcolumntype{L}[1]{>{\centering\arraybackslash}m{#1cm}}

\makeglossaries
\newacronym{ssl}{SSL}{self-supervised learning}
\newacronym{mri}{MRI}{magnetic resonance imaging}
\newacronym{ct}{CT}{computed tomography}
\newacronym{us}{US}{ultrasound}
\newacronym{cnn}{CNN}{convolutional neural network}
\newacronym{cxr}{CXR}{chest X-ray}
\newacronym{auc}{AUC}{area under the receiver operating characteristic curve}

\begin{document}
\maketitle

\begin{abstract}
Self-supervised pretraining has been observed to be effective at improving feature representations for transfer learning, leveraging large amounts of unlabelled data. This review summarizes recent research into its usage in X-ray, computed tomography, magnetic resonance, and ultrasound imaging, concentrating on studies that compare self-supervised pretraining to fully supervised learning for diagnostic tasks such as classification and segmentation. The most pertinent finding is that self-supervised pretraining generally improves downstream task performance compared to full supervision, most prominently when unlabelled examples greatly outnumber labelled examples. Based on the aggregate evidence, recommendations are provided for practitioners considering using self-supervised learning. Motivated by limitations identified in current research, directions and practices for future study are suggested, such as integrating clinical knowledge with theoretically justified self-supervised learning methods, evaluating on public datasets, growing the modest body of evidence for ultrasound, and characterizing the impact of self-supervised pretraining on generalization.
\end{abstract}


\section{Introduction}
\label{sec:introduction}

Significant advancements in deep computer vision has resulted in a surge of interest in applications to medical imaging. Indeed, an enormous number of publications have demonstrated the capabilities of deep learning methods in approximating diagnostic functions in radiological, histological, microsopic, endoscopic, and dermatological imaging. Deep learning has markedly improved in recent years at diagnostic pattern recognition tasks such as classification, object detection, and segmentation in medical imaging.

Of course, methodological advances alone are insufficient to achieve nontrivial results for deep computer vision tasks. Large labelled datasets are the major precondition for success in supervised learning problems. Fortunately, these exist some notable examples of large, open datasets for medical images that contain expert classification labels for a limited set of conditions (e.g., {\tt CheXpert}~\cite{irvin2019chexpert}). Regrettably, large medical imaging datasets containing task- or pathology-specific labels for all constituent examples are substantially less abundant than those containing natural images. Obstacles such as patient privacy concerns, private corporate interests, and expert labelling cost and availability, hamper the production and dissemination of such datasets. Occasionally, situations arise in which unlabelled datasets of medical images are available. Labelling a complete dataset requires established expertise, the cost of which dwarfs the cost of crowdsourcing labels. Furthermore, tasks such as semantic and instance segmentation require greater attention to detail, significantly increasing the labelling time per example.

\begin{figure}[h!]
    \centering
    \includegraphics[width=\textwidth]{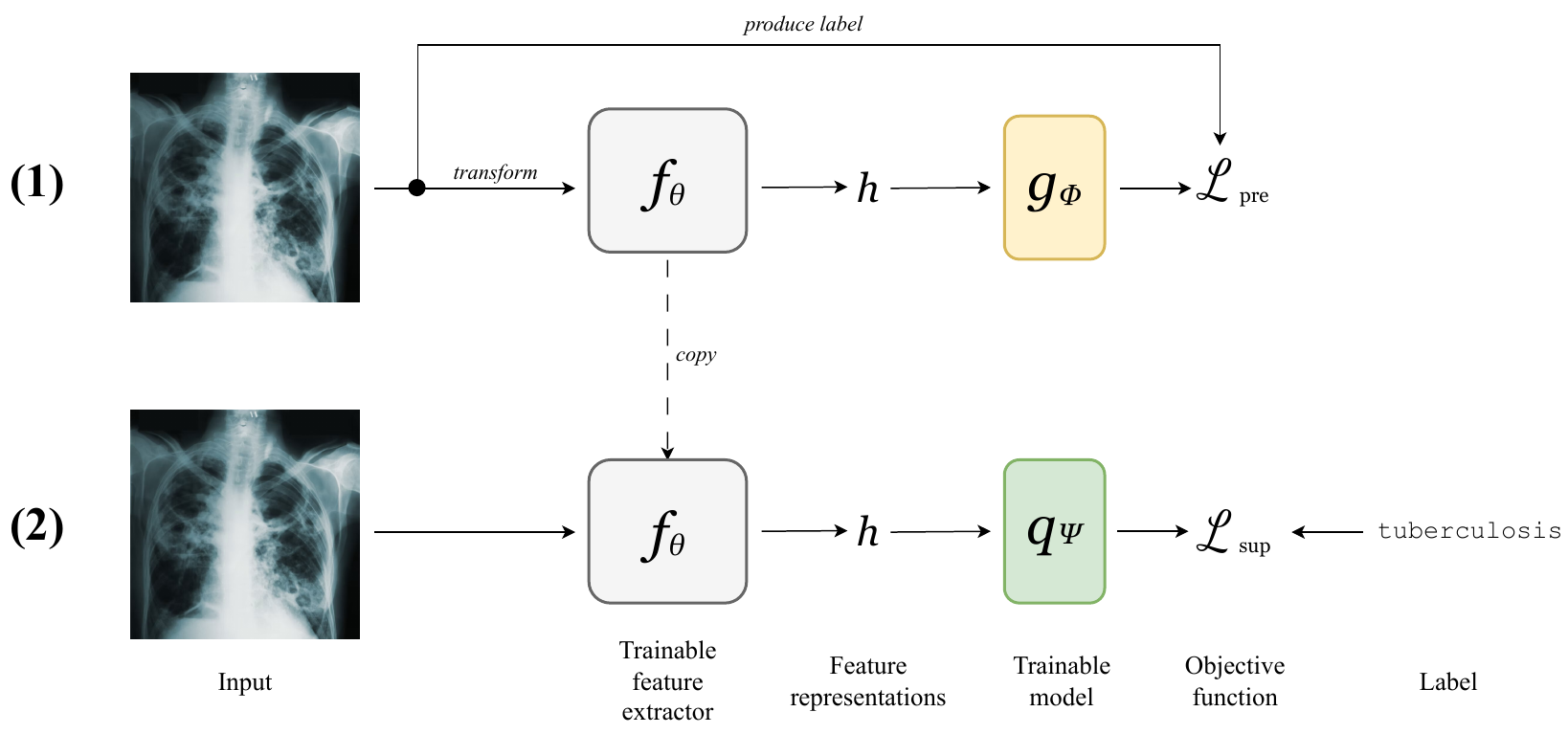}
    \caption{Example of a typical SSL workflow, with an application to chest X-ray classification. \textbf{(1)} \textit{Self-supervised pretraining:} A parameterized model $g_\phi(f_\theta(\mathbf{x}))$ is trained to solve a pretext task using only the chest X-rays. The labels for the pretext task are determined from the inputs themselves, and the model is trained to minimize the pretext objective $\mathcal{L}_{\text{pre}}$. At the end of this step, $f_\theta$ should output useful feature representations. \textbf{(2)} \textit{Supervised fine-tuning:} Parameterized model $q_\psi(f_\theta(\mathbf{x}))$ is trained to solve the supervised learning task of chest X-ray classification using labels specific to the classification task. Note that the previously learned $f_\theta$ is reused for this task, as it produces feature representations specific to chest X-rays.}
    \label{fig:ssl-example}
\end{figure}

\textit{Self-supervised learning} (\acrshort{ssl}) has emerged as a broad strategy to learn a machine learning model that produces feature representations from unlabelled data. It is particularly beneficial when only a subset of examples in a dataset have associated labels. In brief, a machine learning model (typically a deep neural network) is trained to optimize a supervised learning objective in which the targets can be derived from the inputs themselves. In other words, a model is trained to solve a \textit{pretext task}, which is a problem that is solvable using only the inputs and that requires no labels. \textit{Self-supervised pretraining} refers to the optimization of a self-supervised objective to obtain a model capable of producing meaningful feature representations that capture salient information available in the inputs. The learned parameters of the pretrained model may then be used to initialize a new model that can be trained to solve a more specific supervised learning problem for which labelled data is available. Figure~\ref{fig:ssl-example} portrays an example of the steps undertaken to pretrain a model using SSL to learn representations of chest X-rays, prior to training a multiclass chest X-ray classifier.

\begin{figure}[h!]
    \centering
    \begin{subfigure}[b]{0.35\textwidth}
        \includegraphics[width=\linewidth]{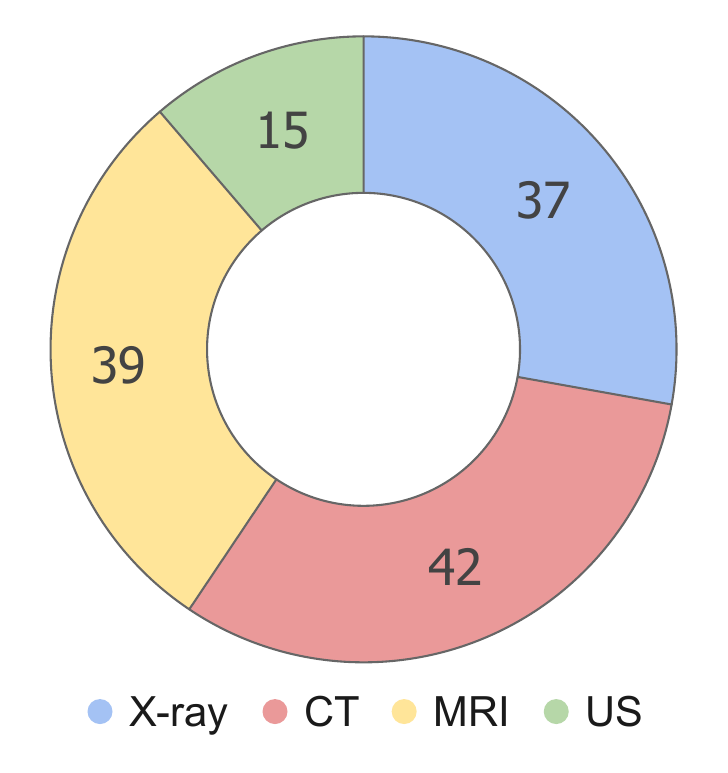}
        \caption{The number of studies presenting experiments for each of the four modalities covered in this review: X-ray, CT, MRI, US. Note that some studies investigated multiple of the above modalities.}
        \label{fig:papers-by-modality}
    \end{subfigure}
    \hfill
    \begin{subfigure}[b]{0.6\textwidth}
        \includegraphics[width=\linewidth]{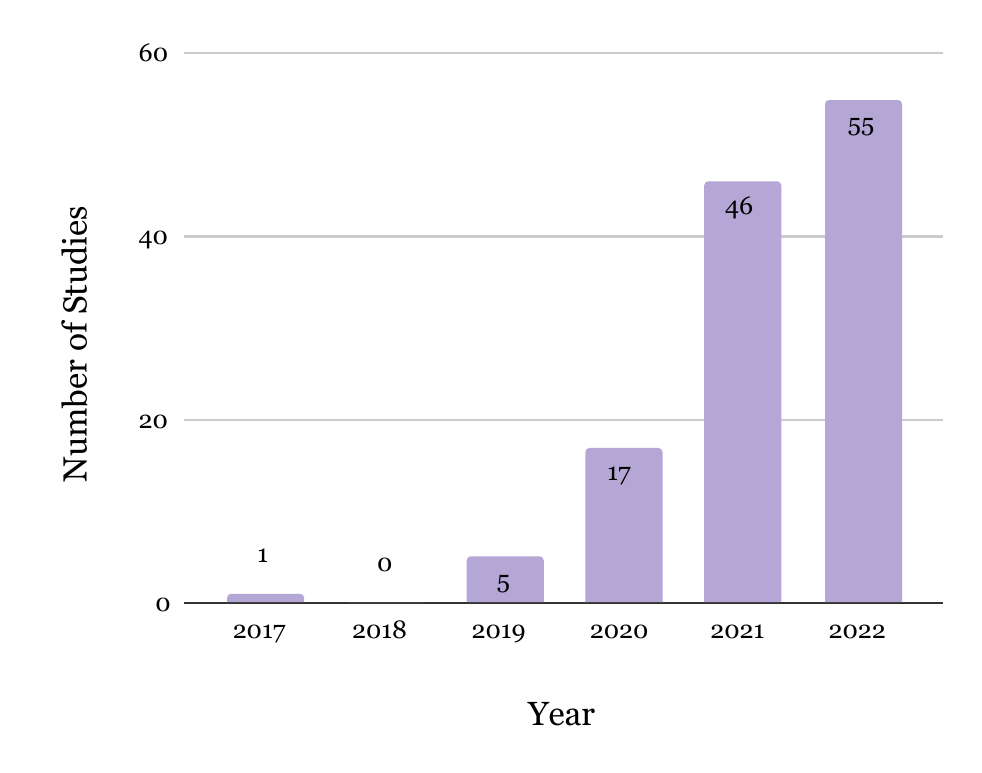}
        \caption{The number of publications by year investigated in this survey, reflecting a growing interest in SSL for radiological imaging.}
        \label{fig:papers-by-year}
    \end{subfigure}
    \caption{Breakdown of the papers included in this survey by (a) imaging modality and (b) year of publication.}
    \label{fig:paper-breakdown}
\end{figure}

SSL is naturally suited to facilitate the advancement of automated diagnostic tasks with radiological images, as vast quantities of historical data are available in picture archiving and communication systems at healthcare institutions worldwide, but labels may not be present. Although accompanying radiological reports may exist in the electronic medical record, it is laborious to devise classification labels from unstructured text. Furthermore, reports may not explicitly identify all relevant negative findings for conditions of interest, opting to omit descriptions of normality. Matters are especially complicated in the context of segmentation tasks. Regardless, it is rare to encounter a fully labelled retrospectively acquired dataset. It is often necessary for experts to label at least a fraction of the dataset. Expert labelling may be prohibitively expensive in terms of monetary cost and/or time. SSL pretraining can therefore materially reduce the burden on experts to label entire radiological datasets.

The purpose of this review is to coalesce and assess evidence that the use of self-supervised pretraining can result in equivalent (and sometimes superior) performance in diagnostic tasks with small fractions of labelled radiological data. Concretely, this review offers the following:

\begin{itemize}
    \item An overview of relevant literature that presents evidence regarding the impact of self-supervised pretraining for diagnostic tasks in radiological imaging, focusing on magnetic resonance imaging (\acrshort{mri}), computed tomography (\acrshort{ct}), radiography (X-ray), and ultrasound (\acrshort{us}).
    \item Identification of areas in the literature that warrant further investigation
    \item Recommendations for future research directions
\end{itemize}

The present work is not the first to review self-supervised approaches in medical imaging. A 2022 survey by Shurrab \& Duwairi~\cite{shurrab_self-supervised_2022} describes common approaches to self-supervised learning and provides examples of studies that have applied it to medical imaging tasks. This review distinguishes itself from~\cite{shurrab_self-supervised_2022} in that it limits its focus to radiological images, excluding other modalities such as histopathological, dermatologic, and endoscopic imagery. The current review goes beyond previous surveys by including more recent studies and addressing deeper theoretical underpinnings.

The remainder of the review is organized in the following manner. First, we describe the literature search methodology that was applied to recover the studies described herein. What follows is an abridged introduction to self-supervised learning. We then present evidence for the merits of SSL as reported by a selection of recent studies -- a separate section is dedicated to each of MRI, CT, X-ray, and US. Prior to the conclusion, we address gaps in the literature and summarize recommendations for future studies.

\section{Search Methodology}
\label{sec:search-methodology}

The purpose of this review is to consolidate and evaluate studies quantifying the benefit of self-supervised pretraining in the automation of diagnostic tasks concerning radiological images. A list of potentially qualifying publications was found by searching the following four databases: Scopus, IEEE, ACM, and PubMed. Queries were designed to cast a wide net, including all studies whose titles, abstracts, keywords, or bodies mention medical images, CT, MRI, X-ray, ultrasound \textit{and} self-supervised learning or contrastive learning. As will be discussed in Section~\ref{sec:ssl}, contrastive learning is a commonly used pretext task in SSL. Appendix A gives the exact queries for each database. The search returned a total of $1226$ results, which was reduced to $778$ unique studies by removing duplicate and completely irrelevant papers.

Exclusion criteria were applied to the results to narrow down the body of literature to those assessing the impact of self-supervised pretraining. Studies were excluded if they were not concerned with radiological imagery, which, for the purposes of this review, include MRI, CT, X-ray, and US examinations. All studies that presented SSL objectives in the context of semi-supervised learning were excluded. Preprints were excluded. Additionally, any study that applied self-supervised learning for a diagnostic task but did not compare performance on their downstream supervised learning task with a baseline (e.g., weights trained from scratch or initialized using weights pretrained on {\tt ImageNet}~\cite{deng2009}). The result was a collection of $124$ studies. Figure~\ref{fig:papers-by-modality} visualizes the distribution of these papers by imaging modality. As shown in the figure, there are considerably less self-supervised pretraining publications geared toward ultrasound tasks than for X-ray, CT, or MRI. Figure~\ref{fig:papers-by-year} compares the number of papers in this survey published per year, reflecting the increasing interest and progress in SSL over the last couple years.

In this review, we directly address all included studies. Given the focused nature of this survey, we focus on studies that concentrate on common downstream tasks. Additionally, we attempt to highlight studies featuring replicable methods, as indicated by evaluation on public datasets and open source availability of experiment code.

\begin{figure}[h!]
    \centering
    \begin{subfigure}[b]{0.35\textwidth}
        \includegraphics[width=\linewidth]{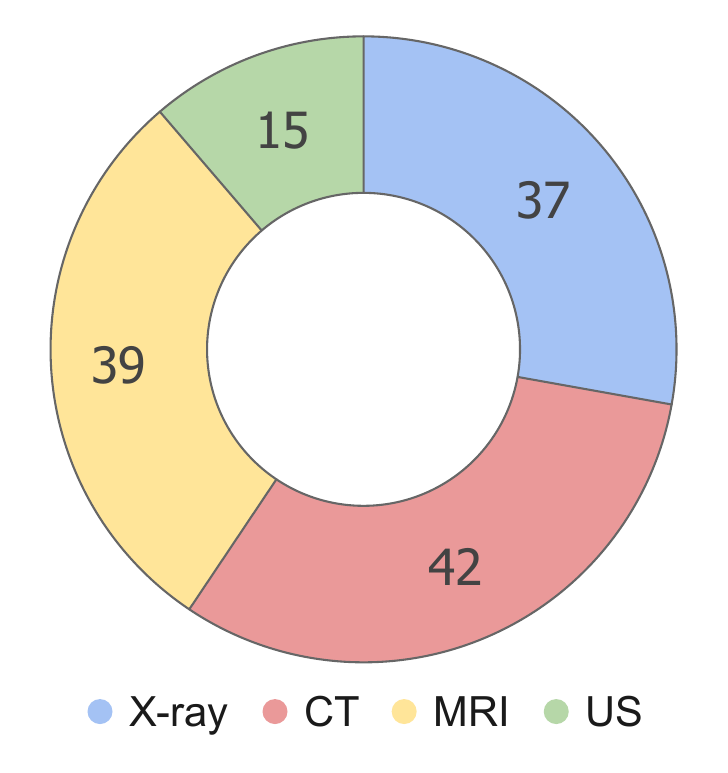}
        \caption{The number of studies presenting experiments for each of the four modalities covered in this review: X-ray, CT, MRI, US. Note that some studies investigated multiple of the above modalities.}
        \label{fig:papers-by-modality}
    \end{subfigure}
    \hfill
    \begin{subfigure}[b]{0.6\textwidth}
        \includegraphics[width=\linewidth]{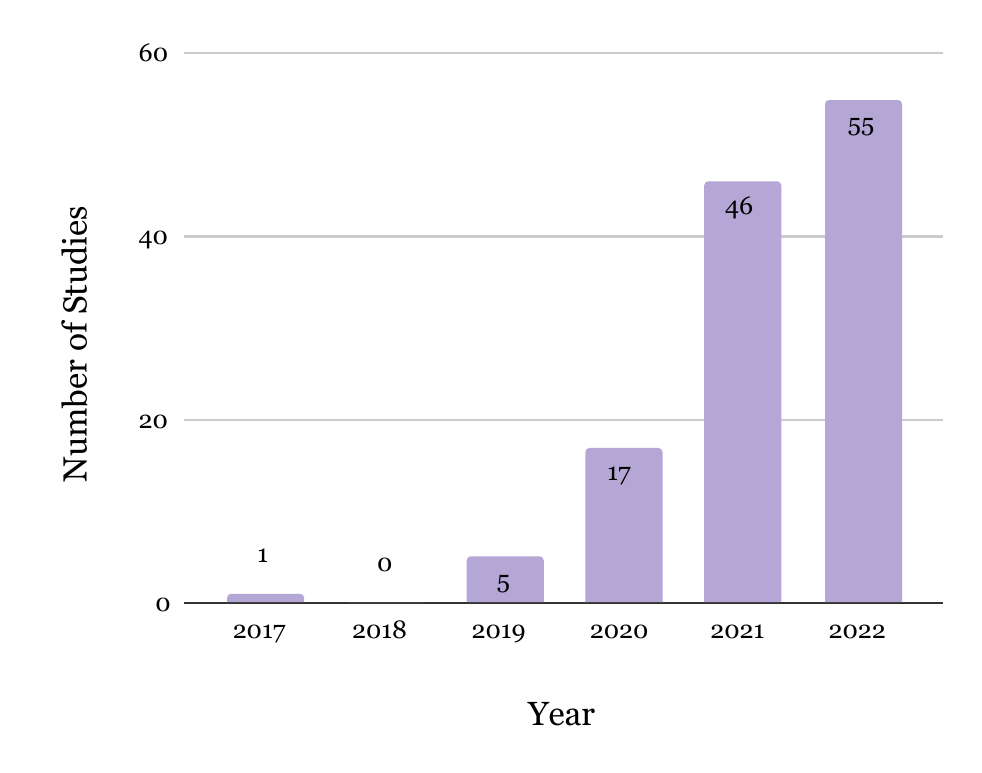}
        \caption{The number of publications by year investigated in this survey, reflecting a growing interest in SSL for radiological imaging.}
        \label{fig:papers-by-year}
    \end{subfigure}
    \caption{Breakdown of the papers included in this survey by (a) imaging modality and (b) year of publication.}
    \label{fig:paper-breakdown}
\end{figure}

\section{Background}
\label{sec:ssl}

\subsection{Preliminaries}
\label{subsec:preliminaries}

In {representation learning}, machine learning models are trained to produce compact $d$-dimensional representations of inputs that are useful for some task(s). SSL is a form of representation learning in which the objective function is formed from a pretext task whose solutions are easily obtainable from unlabelled examples. SSL distinguishes itself from supervised learning in that the objective does not depend on labels for some specific task. Like unsupervised learning, SSL aims to derive compact, low-dimensional representations for examples; however, it is distinct in that it involves optimizing supervised learning objectives. 

The goal of SSL is to learn a feature extractor (i.e., ``backbone") that can extract high-level representations from examples. The weights of the feature extractor may then be applied to subsequent (often referred to as ``downstream") supervised learning tasks for which labels are available. The weights of the feature extractor may be kept stagnant or fine-tuned in the downstream learning problem. To gain intuition into the advantage of learning representations with SSL, consider the following example. Suppose a toddler is seeing different kinds of fruits for the first time. Without any feedback or external knowledge, they discover attributes of the fruits that distinguish them from others, such as colour and shape. Later, when they are taught to identify fruits by name in preschool, they apply their previously acquired knowledge about fruits to help her classify them (e.g., limes are green and round). It is likely that they have an advantage over classmates who did not eat fruit at home.

More concretely, suppose that we have a dataset of examples, $\mathcal{X}$. A {\it pretext task} is formulated that is solvable with knowledge of the examples. Note that the task may be defined for one or more examples. Solutions for the pretext task are taken as the labels for a self-supervised problem. An objective is defined that appropriately measures the performance of a learner at solving the pretext task. 

In the context of computer vision, a backbone model, $f_\theta: \mathbb{R}^{h \times w \times c} \rightarrow \mathbb{R}^d$, is defined that maps $h \times w \times c$ images to a $d$-dimensional representation. $f_\theta$ is typically a deep neural network, parameterized by $\theta$, whose architecture embodies an inductive bias amenable to the equivariance and invariance relationships inherent to the dataset, such as a convolutional neural network (\acrshort{cnn}). The objective is computed from the output of a secondary function $g_\phi: \mathbb{R}^{d} \rightarrow \mathbb{R}^{e}$, where $g_\phi$ is a neural network with parameter $\phi$. The pretext objective $\mathcal{L}_{\text{pre}}$ is then optimized to recover optimal weights $\theta^*$ and $\phi^*$.

\begin{equation}
    (\theta^*, \phi^*) = \arg \min_{\theta, \phi} \mathcal{L}_{\text{pre}}(g_\phi(f_\theta(\mathbf{x})))
\end{equation}

For the chest X-ray classification example in Figure~\ref{fig:ssl-example}, suppose chest X-ray images are passed to a CNN feature extractor $f_\theta$. The resulting feature representations $\mathbf{h}$ are passed to multilayer perceptron $g_\phi$, the output of which is used to compute $\mathcal{L}_{\text{pre}}$, which quantifies performance on the pretext task.

After the objective is optimized, $g_\phi$ is customarily discarded. The backbone $f_\theta$ may then be applied for a subsequent supervised learning problem, as $f_\theta(\mathbf{x})$ is a nontrivial representation of $\mathbf{x}$. For a supervised learning task with examples $\mathcal{X}'$ (originating from an identical or similar distribution as $\mathcal{X}$) and corresponding labels $\mathcal{Y}$, a new model head $q_\psi: \mathbb{R}^d \rightarrow \mathbb{R}^{\text{dim}(y)}$ is initialized. $q_\psi$ receives feature representations $\mathbf{h}$ as input. The model $q_\psi(f_\theta(\mathbf{x}))$ is trained to minimize a loss function with respect to the labels. At this stage, $\theta$ may be held constant or fine-tuned via transfer learning. Broadly, this process is referred to as {\it self-supervised pretraining}. Note that it is possible that the pretrained weights $\theta$ may constitute a useful initialization for multiple downstream supervised learning problems. 

\subsection{SSL Approaches}
\label{subsec:approaches}

The major difference between various self-supervised pretraining methods is the choice of pretext task and its optimization. Here we enumerate some broad categories of SSL methods. The intention is to provide the reader with a high-level understanding of the main approaches to SSL that may be useful when describing specific studies in the subsequent sections. These approaches are often trialled on natural images first, likely due to the high availability of benchmark datasets and broad applicability.

\subsubsection{Generative Methods}

\begin{figure}[h!]
    \centering
    \begin{subfigure}[b]{0.46\textwidth}
        \includegraphics[width=\linewidth]{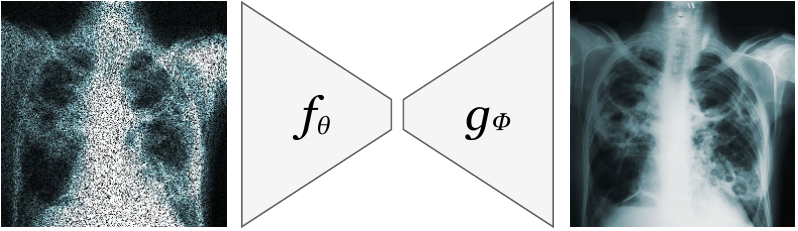}
        \caption{Denoising~\cite{vincent2008extracting}}
        \label{fig:denoising_autoencoder}
    \end{subfigure}
    \hfill
    \begin{subfigure}[b]{0.46\textwidth}
        \includegraphics[width=\linewidth]{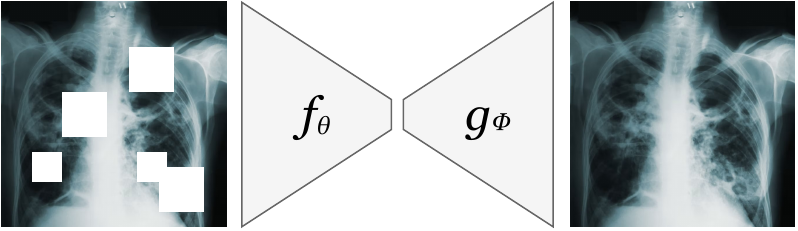}
        \caption{Inpainting~\cite{pathak2016context}}
        \label{fig:inpainting}
    \end{subfigure}
    \caption{examples of generative SSL pretext tasks}
    \label{fig:generative_methods}
\end{figure}

Several SSL pretext tasks are built around generative modelling. The output of $g_\phi$ is an entire image or a fragment of an image. Generative methods often employ an \textit{encoder} that learns rich feature representations. The feature representations are sent to a secondary network, frequently referred to as a \textit{decoder}. In a self-supervised context, $g_\phi$ is the decoder and $f_\theta$ is the encoder, which is retained for downstream supervised learning. Many generative tasks are reconstructive, in that they recover a corrupted version of an image. An example of a reconstructive approach to self-supervised learning is the denoising autoencoder~\cite{vincent2008extracting} (Figure~\ref{fig:denoising_autoencoder}). In the image colourization task, coloured images are generated from greyscale images, which is made possible by the availability of a dataset of coloured images~\cite{zhang2016colorful}. Inpainting of redacted patches of images is another example of a reconstructive pretext task~\cite{pathak2016context} (Figure~\ref{fig:inpainting}).

\subsubsection{Predictive Methods}

\begin{figure}[h!]
    \centering
    \begin{subfigure}[b]{0.33\textwidth}
        \includegraphics[width=\linewidth]{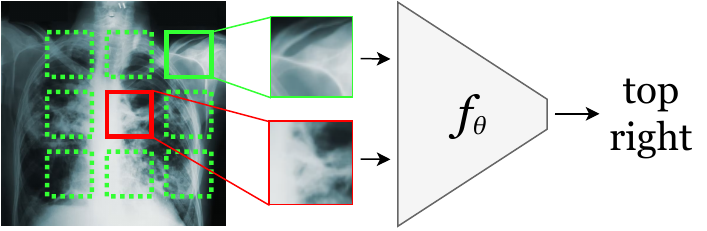}
        \caption{Context prediction~\cite{doersch2015unsupervised}}
        \label{fig:patch_location}
    \end{subfigure}
    \hfill
    \begin{subfigure}[b]{0.33\textwidth}
        \includegraphics[width=\linewidth]{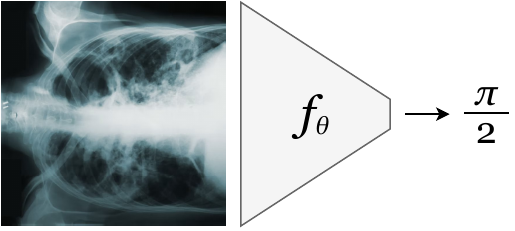}
        \caption{Rotation prediction~\cite{gidaris2018unsupervised}}
        \label{fig:rotation}
    \end{subfigure}
    \hfill
    \begin{subfigure}[b]{0.33\textwidth}
        \includegraphics[width=\linewidth]{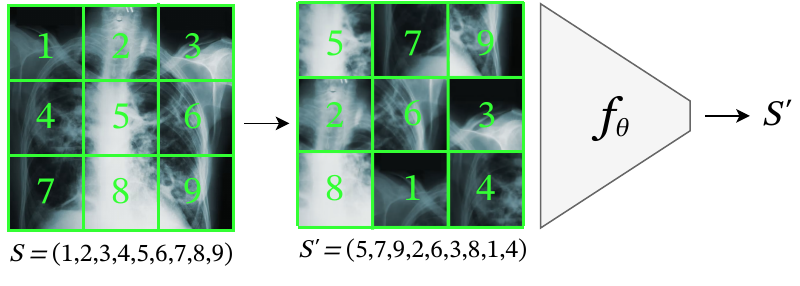}
        \caption{Patch permutation prediction (jigsaw)~\cite{noroozi2016unsupervised}}
        \label{fig:jigsaw}
    \end{subfigure}
    \caption{Examples of predictive SSL pretext tasks}
    \label{fig:predictive_methods}
\end{figure}

Many custom pretext tasks have been proposed for computer vision that involve learning a specific transformation applied to images. A stochastic transformation is applied to each example, and the learner's task is to predict or to undo the transformation. For instance, the context prediction task is defined as the problem of predicting the relative location of random image patches from unlabelled images (Figure~\ref{fig:patch_location})~\cite{doersch2015unsupervised}. In the rotation prediction task, a random rotation is applied to an image and the learner must infer which rotation was applied (Figure~\ref{fig:rotation})~\cite{gidaris2018unsupervised}. The jigsaw task is the unscrambling of a random permutation of all the rectangular patches in an image (Figure~\ref{fig:jigsaw})~\cite{noroozi2016unsupervised}. Generally, the label is defined as the transformation that was applied to the image. The transformation may be stochastic in that its parameters may be sampled from some underlying distribution (e.g., the angle of a rotation being sampled from a multinoulli distribution over predefined angles). 

A criticism of transformation prediction methods is that they may not be generally applicable to downstream tasks because the pretext tasks are formulated using specialized heuristics~\cite{chen2020simple}. \textit{Contrastive learning} has evolved as a generic approach for learning feature representations with fewer assumptions regarding the usefulness of particular tasks. Framed succinctly, the contrastive learning problem is to produce representations that are invariant to non-meaningful transformations. In contrastive learning, $g_\phi(\mathbf{h})$ is a neural network that outputs vector \textit{embeddings}, which need not have the same dimension as the representations $\mathbf{h}$. The goal of contrastive learning is to produce embeddings $\mathbf{z}_i$ that are very close (as measured by some distance function, $d(\mathbf{z}_i, \mathbf{z}_j)$) for \textit{positive pairs} of examples and very far for \textit{negative pairs}. SimCLR~\cite{chen2020simple} is an example of a contrastive learning SSL method in which positive pairs are distorted versions of the same image and negative pairs are distorted versions of distinct images. The weights $\theta$ and $\phi$ are optimized such that the embeddings are close and far for positive and negative pairs respectively. To produce distorted versions of images, a series of data augmentation transformations is applied, where the parameters of the transformation are sampled from a probability distribution. Common examples of transformations include affine transformations, noise addition, and adjustments to brightness, contrast, and hue. Other notable examples of contrastive learning in SSL include MoCo~\cite{he2020momentum}, and PIRL~\cite{mishra_cr-ssl_2022}.

\begin{figure}[h!]
    \centering
    \includegraphics[width=\linewidth]{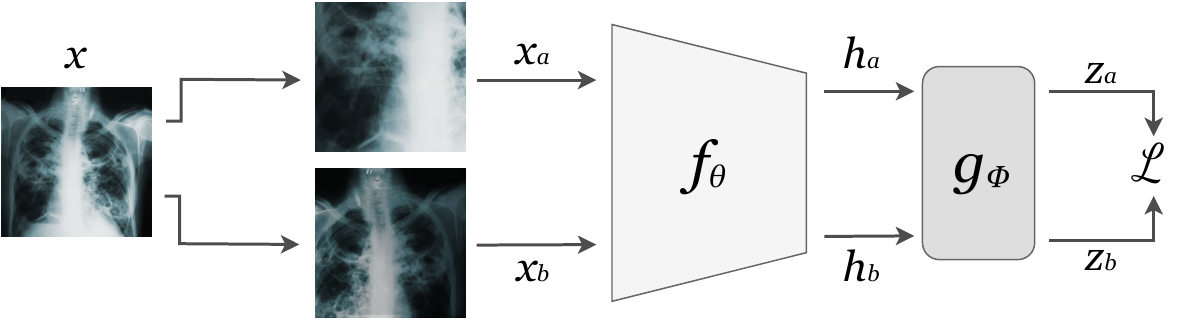}
    \caption{A depiction of the forward pass for a positive pair in a standard noncontrastive pretext task. An image is subject to stochastic data transformations twice, producing distorted views $\mathbf{x}_a$ and $\mathbf{x}_b$, which are passed through the feature extractor $f_\theta$ to yield feature representations $\mathbf{h}_a$ and $\mathbf{h}_b$. The projector $g_\phi$ transforms $\mathbf{h}_a$ and $\mathbf{h}_b$ into embeddings $\mathbf{z}_a$ and $\mathbf{z}_b$ respectively. Typically, the objective $\mathcal{L}$ is optimized to maximize the similarity of $\mathbf{z}_a$ and $\mathbf{z}_b$.}
    \label{fig:noncontrastive_learning}
\end{figure}

A major obstacle in contemporary contrastive learning approaches is the reliance on vast quantities of negative pairs, necessitating large batch sizes~\cite{zbontar2021barlow}. Several recent publications have focused on approaches relying only on positive pairs, collectively referred to as \textit{noncontrastive learning} (Figure~\ref{fig:noncontrastive_learning}). Different transformations are applied to the same image to produce multiple views. $f_\theta$ and $g_\phi$ are optimized to produce embeddings that are robust to the possible views entailed by the transformation distribution, through the minimization of distance between the embeddings of positive pairs. Various strategies have been devised to avoid the problem of \textit{information collapse}, where models learn the trivial solution of indiscriminately predicting embeddings zero vectors. Examples of methods that have reported results comparable or superior to contrastive learning include BYOL~\cite{grill2020bootstrap}, Barlow Twins~\cite{zbontar2021barlow}, and VICReg~\cite{bardes2022vicreg}. 

\subsection{Theoretical Support}
\label{subsec:theoretical-justification}

Until recently, SSL publications were focused primarily on introducing novel methods guided by intuitions. Some researchers have since attempted to explore the properties of SSL pretraining to better understand why they deliver such benefits and to ascertain conditions under which they will succeed.

Efforts in attempting to understanding the efficacy of optimizing performance on pretext tasks in learning downstream tasks are growing. Lee \etal~\cite{lee2021predicting} provided guarantees for the improved sample efficiency of pretraining  with a reconstructive pretext task, in scenarios where the inputs and pretext target are conditionally independent of the labels and a latent variable. Dropping the conditional independence assumption, HaoChen \etal~\cite{haochen2021provable} defined a contrastive loss based on spectral decomposition and derived performance guarantees for linear classifiers trained on the feature representations from the pretraining phase. Most recently, Balestriero \& LeCun~\cite{balestriero2022contrastive} developed an amalgamated lens through which contemporary contrastive and noncontrastive approaches may be viewed, based on spectral analysis. They demonstrated that a selection of SSL methods (including Barlow Twins~\cite{zbontar2021barlow}, VICReg~\cite{bardes2022vicreg}, and SimCLR~\cite{chen2020simple}) are optimal choices for solving downstream tasks as long as the relation between labels is included in the relationship between positive pairs~\cite{balestriero2022contrastive}. Practitioners in all domains of computer vision are therefore encouraged to ensure that their choice of pretext task aligns appropriately with the label distribution. Those applying SSL pretraining for radiological imaging tasks should consider these results when selecting a pretext task.

\section{Applications In Radiograph Imaging}

The medical imaging machine learning community has extensively reported on automatic interpretation of radiographs (X-rays). A large fraction of the effort has focused on interpretation of chest X-rays (\acrshort{cxr}). There exists an overwhelming volume of literature describing the use of deep neural networks for CXR classification tasks. A major enabling force for this work has been the availability of large, publicly available, labelled datasets. Perhaps unsurprisingly, a flurry of studies exploring the use of self-supervised pretraining for CXR analysis followed. Despite the prevalence of open datasets, it remains difficult to directly compare publications, since pretraining and evaluation protocols differ. Here we summarize the results of such publications to understand the impact of SSL.

\subsection{Chest X-ray Diagnostic Tasks}

Evidence has been presented in favour of self-supervised pretraining for chest X-rays, with reported benefits ranging from improved performance, label efficiency, and robustness to external data distributions. Many studies focus on the problem of identifying common respiratory conditions in CXR for which labels are available in large public datasets. A substantial fraction of publications focus on identifying COVID-19 in CXR, which is likely due to the co-occurrence of the COVID-19 pandemic and the escalation of SSL popularity.

Contrastive learning approaches have been extensively studied in the context of CXR classification. In 2020, Zhou \etal~\cite{zhou_comparing_2020} introduced C2L, a joint embedding contrastive learning approach that employs a batch-wise mixup operation and a teacher network with momentum updates. Pretraining was conducted on a constellation of publicly available datasets. When fine-tuning and evaluating on {\tt Chest X-ray14}~\cite{wang2017chestxray14}, {\tt CheXpert}~\cite{irvin2019chexpert}, and {\tt RSNA Pneumonia}~\cite{rsna-pneumonia-detection-challenge}, C2L outperforms supervised models pretrained on {\tt ImageNet} and self-supervised models pretrained with MoCo~\cite{he2020momentum}. Other variants of MoCo have also exhibited improvement over fully supervised learning for CXR classification~\cite{dong_federated_2021, liao_muscle_2022, dalla_serra_improving_2022, ridzuan_self-supervision_2022}. Table~\ref{tab:chestxray14-compare} provides average class-wise area under the receiver operating characteristic curve (\acrshort{auc}) reported by multiple studies for the official {\tt Chest X-ray14} test set after pretraining and training on the training set.

\begin{table}[h!]
    \centering
    \begin{tabular}{cccccc}
         \toprule
         \multicolumn{3}{c}{{\sc Method}} &  \multicolumn{3}{c}{{\sc Initialization}} \\
         \cmidrule(lr){1-3} \cmidrule(lr){4-6}
         \small
         First Author [ref] & Identifier & Extractor & Random & {\tt ImageNet} & SSL \\
         \midrule
         Zhou \cite{zhou_comparing_2020} & C2L & ResNet18 & - & $0.8150$ & $0.8350$ \\
         Zhou \cite{zhou_comparing_2020} & C2L & DenseNet121 & - & $0.8290$ & $0.8440$ \\
          Ma \cite{ma_benchmarking_2022} & SimMIM & ViT-B & $0.7169$ & - & $0.7955$ \\ 
         Ma \cite{ma_benchmarking_2022} & SimMIM & Swin-B & $0.7704$ & - & $0.8195$ \\ 
         Liu  \cite{liu_self-supervised_2021} & S\textsuperscript{2}MTS\textsuperscript{2} & DenseNet121 & - & - & $0.8250$ \\
         Haghighi  \cite{haghighi_dira_2022} & DiRA & ResNet-50 & $0.8031$ & $0.8170$ & $0.8112$ \\
         Pang  \cite{pang_popar_2022} & POPAR & Swin-B & $0.7429$ & $0.8132$ & $0.8181$ \\
         \bottomrule
    \end{tabular}
    \caption{A comparison of SSL pretraining studies that investigated chest X-ray classification using the {\tt Chest X-ray14} dataset for pretraining, training, and testing using the official splits. The table gives class-wise average test AUC as reported by the authors, when training using all available labels.}
    \label{tab:chestxray14-compare}
\end{table}

Azizi \etal~\cite{azizi_big_2021} conducted a thorough study into the efficacy of SSL pretraining for CXR classification using a variant of SimCLR~\cite{chen2020simple}, reporting improvments in mean AUC of over $0.01$ when pretrained on {\tt CheXpert}, as compared to fully supervised models. The authors' approach, named Multi-Instance Contrastive Learning, generalizes positive pairs to include CXRs of the same patient case, thereby exploiting information already available in the dataset to complicate the pretext task. Valuable insights were derived from their investigations. Notably, the authors found that the best-performing strategy was to initialize the weights of feature extractors with {\tt ImageNet}-pretrained weights prior to conducting pretraining. Experiments also established that SSL-pretrained models outperformed fully supervised models when evaluated on {\tt Chest X-ray14}, an external dataset. Other studies have reported that SimCLR pretraining (and variants) yield improvements in CXR classification~\cite{zhao_fast_2021, feki_self-supervised_2022} and CXR object detection~\cite{bencevic_self-supervised_2022}. 
Several other publications report improvements in performance on downstream tasks using customized contrastive learning approaches for CXR diagnostic tasks~\cite{han_pneumonia_2021, gazda_self-supervised_2021, liu_self-supervised_2021, konwer2022clinical, hao_self-supervised_2022, li_self-knowledge_2022, wei_triplet_2022}

Noncontrastive approaches have also been explored for CXR diagnostic tasks. Nguyen \etal~\cite{nguyen_semi-supervised_2021} applied BYOL to pretrain CXR classifiers using the {\tt ChestMNIST} and {\tt PneumoniaMNIST} datasets (originating from {\tt MedMNIST}~\cite{medmnistv2}, achieving significantly higher AUC on the downstream binary classification tasks than supervised models initialized randomly or with {\tt ImageNet}. Mondal \etal~\cite{mondal_covid-19_2022} also witnessed improvement of COVID-19 classification on the {\tt COVIDx CXR-2} dataset~\cite{zhao2021covidx} when pretraining on {\tt CheXpert} using BYOL.

Alternative pretext tasks have also yielded improvements in CXR tasks. Pang \etal~\cite{pang_popar_2022} described the use of patch de-shuffling and recovery for pretraining vision transformers, demonstrating superior performance compared against fully supervised learning alone. Ma \etal~\cite{ma_benchmarking_2022} demonstrated the benefit of masked image modelling for pretraining vision transformers for various CXR tasks. Haghighi \etal~\cite{haghighi_dira_2022} proposed \textit{DiRA}, which combines discriminative methods (namely, SimSiam\cite{chen2021exploring}, MoCo~\cite{he2020momentum}, and Barlow Twins~\cite{zbontar2021barlow}), restoration of distorted images, and adversarial training into a composite pretext task. Improvements over supervised training were not observed when fine-tuning on the same dataset that was used for pretraining; however, statistically significant improvements were noted when the pretraining dataset did not match the dataset in the downstream task. Interestingly, the method outperforms each of SimSiam, MoCo, and Barlow Twins alone, indicating the possible value of composite pretext tasks. Other pretext tasks investigated for CXR classification include reconstruction of original images after transformation~\cite{zhou_preservational_2021} or distortions and/or masking~\cite{park_deep_2021}, data augmentation prediction~\cite{tang_enhancing_2021}, and predicting pseudo-labels generated using sample decomposition~\cite{abbas_4s-dt_2021}.

Multi-modal pretext tasks have also been explored for CXR analysis. Some public CXR datasets contain accompanying textual reports, which can be exploited to produce rich feature representations that align with physician impressions. For instance, Ji \etal~\cite{ji_improving_2021} pretrained a network that learns similar representations for paired CXRs and reports. Müller \etal~\cite{muller_radiological_2022} demonstrate that contrastive pretraining that maximizes similarity between CXRs and reports improves performance on downstream CXR object detection and segmentation tasks on multiple public datasets, compared with image-only pretraining or fully supervised learning. In a similar approach, Tiu \etal~\cite{tiu_expert-level_2022} maximized the cosine similarity of paired CXR images and ``Impressions" sectionss of reports from the {\tt MIMIC-CXR} dataset. In lieu of fine-tuning, the authors evaluated the trained vision transformer by providing textual prompts containing the label and taking the maximum of the logits (e.g., ``pneumothorax" and ``no pneumothorax") to determine the presence or absence of conditions. This zero-shot learning approach nearly matched fully supervised approaches' performance. The success of multi-modal schemes is made possible by datasets where physician reports accompany images, such as {\tt MIMIC-CXR}~\cite{johnson2019mimic}.

\subsection{Breast Cancer Identification}

Another major diagnostic task for which many deep learning solutions have been proposed is the identification of anomalies seen on mammograms that could be cancerous. Truong \etal~\cite{truong_vu_improved_2021} observed that pretraining to solve the jigsaw pretext task improved prediction of malignant breast lesions when only a quarter of labels are available. You \etal~\cite{you_intra-class_2022} demonstrated that a contrastive learning pretext task outperforms the baseline. The pretext task was unique in that it considered multiple views of the same breast as positive pairs. Treating bilateral mammograms as a positive pair improved the performance of a breast cancer screening model~\cite{cao_supervised_2021}. Finally, BYOL~\cite{grill2020bootstrap} pretraining was shown to improve breast tumour segmentation~\cite{saidnassim_self-supervised_2021}. In contrast with CXR studies, mammogram studies have no publicly available data, limiting the replicability of their results.

\subsection{Oral Radiographs}

Taleb \etal~\cite{taleb_self-supervised_2022} investigated the utility of SimCLR, Barlow Twins, and BYOL for pretraining a CNN to detect dental caries, boosting sensitivity by up to $6\%$ and outperforming humans when fine-tuned using only $152$ images. Hu \etal observed that pretraining using a reconstruction pretext task improves downstream classification and segmentation of jaw tumours and cysts.

\section{Applications in Computed Tomography}

Deep computer vision has been heavily drawn upon for automated CT analysis. CT scans are volumetric scans; as a result, 3D CNNs are often leveraged. 2D CNNs are also applied for problems where a single saggital, coronal, or axial image is sufficient for the target task. Vision transformers are increasingly being studied as well. Segmentation of organs and lesions are common examples of machine learning tasks in CT. There are two major types of segmentation tasks: \textit{semantic segmentation} consists of labelling each pixel in an image according to the class to which it belongs, and \textit{instance segmentation} involves identifying distinct objects in an image and designating its consituent pixels. Semantic and instance segmentation tasks require greatly increased labelling time compared to classification tasks. Evidence for improved label efficiency resulting from SSL pretraining underlines its value as a cost reduction strategy. In this section, we explore the impact of SSL pretraining for CT, categorized by application.

\subsection{Lung Nodule Detection \& Segmentation}

The {\tt LIDC-IDRI} database is a large, labelled, public collection of CT scans with lung nodule annotations and segmentation masks~\cite{armato2011LIDC-IDRI}. It comprised the dataset for the {\tt LUNA2016} challenge~\cite{setio2017luna2016}, which was an open competition aimed at finding machine learning solutions to lung cancer screening. It became a common CT computer vision benchmark, and many SSL studies have utilized it.

Referenced by multiple succeeding publications, Models Genesis~\cite{zhou_models_2021} devised a restorative approach for pretraining on subvolumes of 3D medical images. The approach involved applying transformations such as nonlinear translation, pixel shuffling, cropping, and masking to the subvolume. An encoder-decoder CNN is pretrained to restore the subvolumes. The encoder is reused for downstream classification tasks, while the entire pretrained encoder-decoder is used for downstream segmentation tasks. The pretrained models are available upon request, strengthening the replicability of their results. Building on Models Genesis, the Semantic Genesis~\cite{haghighi_learning_2020} approach adds a classification loss to the reconstruction loss. The classification task is to predict which class a subregion belongs to, where the classes are constructed for clusters in the latent space of a pretrained autoencoder, which the authors claim contain rich semantic features. The Parts2Whole~\cite{feng_parts2whole_2020} pretext task involves reconstructing a CT volume from a random subvolume. Since the above three methods were tested on the same {\tt LUNDA2016} splits, they can be directly compared (see Table~\ref{tab:ncc-comparison}). For 3D volume inputs, the above methods are superior to training a lung nodule detector from scratch. However, 2D slice-based models pretrained with Models Genesis or Semantic Genesis do not outperform fully supervised models initialized with {\tt ImageNet}-pretrained weights. Other SSL approaches that report improvement over training from scratch have been reported for this problem, but are not directly comparable due to having been trained and/or evaluated on different subsets of {\tt LIDC-IDRI}~\cite{zhai_mvcnet_2022, huang_self-supervised_2022, niu_unsupervised_2022, gai_using_2022, guo_discriminative_2022}. Several of the aforementioned studies have also observed increase in performance for lung nodule segmentation on \cite{armato2011LIDC-IDRI} when pretraining using their own or preceding SSL methods~\cite{zhou_models_2021, haghighi_learning_2020, feng_parts2whole_2020, guo_discriminative_2022}.

\begin{table}[h!]
    \centering
    \begin{tabular}{ccccc}
         \toprule
         \multicolumn{2}{c}{{\sc Method}} &  \multicolumn{3}{c}{{\sc Initialization}} \\
         \cmidrule(lr){1-2} \cmidrule(lr){3-5}
         \small
         First Author [ref] & Identifier &  Random & {\tt ImageNet} & SSL \\
         \midrule
        \multirow{2}{*}{Zhou \cite{zhou_models_2021}} & Models Genesis (2D) & $0.9603$ & $0.9779$ & $0.9745$ \\ 
          & Models Genesis (3D) & $0.9603$ & N/A & $0.9834$ \\ 
         \multirow{2}{*}{Haghighi \cite{haghighi_learning_2020}} & Semantic Genesis (2D)\tablefootnote{Values were estimated via visual inspection, since results were reported in a chart.} & $0.9425$ & $0.9750$ & $0.9750$ \\
          & Semantic Genesis (3D) & $0.9425$ & N/A & $0.9847$ \\ 
         Feng \cite{feng_parts2whole_2020} & Parts2Whole (3D) & $0.9425$ & N/A & $0.9867$ \\
         \bottomrule
    \end{tabular}
    \caption{A comparison of SSL pretraining studies for 2D and 3D CNNs that investigated lung nodule detection using the {\tt LIDC-IDRI} dataset and the {\tt LUNA 2016} splits. The table gives test AUC as reported by the authors. }
    \label{tab:ncc-comparison}
\end{table}

\subsection{Pulmonary Embolism Detection \& Segmentation}
\label{subsec:pe-detection}

Constructed from the private dataset used by Tajbakhsh \etal~\cite{tajbakhsh2015PE-CAD}, {\tt ECC} is a private benchmark that contains chest CT scans, along with labels that differentiate true pulmonary emboli from false positives. Models Genesis~\cite{zhou_models_2021} and Parts2Whole~\cite{feng_parts2whole_2020} both report a substantial improvement over training 3-dimensional (3D) CNNs from scratch, with Models Genesis achieving slightly higher test AUC than Parts2Whole in a direct comparison. Once again, 2D CNNs pretrained with Models Genesis outperform training from scratch, but do not clearly outperform models initialized with {\tt ImageNet}-pretrained weights. Redesigning conventional discriminative pretext tasks (e.g., jigsaw, rotation) to include reconstructive and adversarial regularizers, Guo \etal~\cite{guo_discriminative_2022} observe consistent improvement on {\tt ECC} using all pretext tasks. It is unclear how the authors of the above studies procured the original private dataset first used in~\cite{tajbakhsh2015PE-CAD}.

{\tt RSNA-PE} is a public dataset containing pulmonary embolism labels for chest CT examinations. Islam \etal~\cite{islam_seeking_2021} pretrained various 2D CNN architectures on ImageNet using an assortment of previously proposed SSL methods, finding that downstream performance on the {\tt RSNA-PE} test set improved for half of the self-supervised methods studied, compared to initialization with {\tt ImageNet}-pretrained weights. Their mixed results are unsurprising, considering that they did not pretrain using CT data. Ma \etal pretrained vision transformers on {\tt RSNA-PE} using SimMIM~\cite{xie2022simmim}, a masked image modelling pretext task, observing a statistically significant improvement in test AUC.

\subsection{Cerebral Hemorrhage Detection}
\label{subsec:cerebral-hem}

An assortment of CT classification tasks have benefitted from self-supervised pretraining. Zhuang \etal~\cite{zhuang_self-supervised_2019} trained a 3D CNN classifier to detect cerebral hemorrhage, applying a custom pretext task they playfully liken to solving a Rubik's cube. The pretext task was to predict the random permutation and rotation applied to the $8$ subvolumes of the cuboid input. Their custom pretraining resulted in $11.2\%$ higher accuracy than training from scratch. Subsequent work modified the task by randomly masking subvolumes, adding a prediction head to classify the masking pattern applied~\cite{zhu_rubiks_2020}. The change resulted in a $1\%$ improvement in accuracy over their previous study. However, the accuracy is lower than in the first study, raising the question of whether the same train/test partitions were applied. Further building on this work, Zhu~\etal~\cite{zhu_aggregative_2022} form an aggregative pretext task that solves multiple proxy tasks, including their prior Rubik's cube method. The pretext tasks are iteratively added after evaluating fine-tuning experiments, and it is unclear if the authors refer to test or training performance. They report improvements over training from scratch using all proxy tasks studied, the greatest accuracy boost being $17.22\%$.

\subsection{COVID-19 Diagnosis}

As in CXR applications, there exist multiple applications to COVID-19 diagnosis in CT. Early in the pandemic, Li \etal~\cite{li_efficient_2020} extend their previous work (Rubik's cube, introduced above in Subsection~\ref{subsec:cerebral-hem}) by randomly masking subvolumes and predicting the mask. It is unclear how this method differs from the masking task delineated in~\cite{zhu_rubiks_2020} -- in fact, the paragraphs describing the masking pretext task are nearly identical in~\cite{zhu_rubiks_2020}~and~\cite{li_efficient_2020}. The authors report an increase in precision but decrease in recall, compared to training from scratch. Interestingly, Ewen \& Khan~\cite{ewen_targeted_2021} achieve better performance on the public {\tt COVID-CT} dataset~\cite{zhao2020covid-ct} by employing a seemingly trivial pretext task of predicting whether a CT scan has been horizontally reflected across the saggital plane. Lu \& Dai~\cite{lu_self-supervised_2022} conducted two rounds of contrastive pretraining using MoCo -- one on the {\tt LUNA2016}~\cite{setio2017luna2016} lung nodule analysis challenge dataset and a second on an expanded version of {\tt COVID-CT}. When evaluating on the {\tt COVID-CT} test set, they observed performance improvement compared to ImageNet pretraining. Hochberg \etal~\cite{hochberg_self_2022} pretrained a StyleGAN and used the convolutional discriminator to initialize a CNN for fine-tuning, observing an improvement over both training from scratch and pretraining with MoCo for COVID-19 detection. Focusing instead on vision transformers, Gai \etal~\cite{gai_using_2022} found that pretraining with DINO~\cite{caron2021dino} substantially improved the AUC of a COVID-19 classifier on the public {\tt COVID-CTset}~\cite{rahimzadeh2021covid-ctset} dataset.

Moving beyond classification, Gao \etal~\cite{gao_denoising_2021} found that pretraining using reconstruction or denoising tasks improved the Dice score of a model trained to segment COVID-19 lesions on CT images. Since their pretext tasks were generative, they were able to use the weights of both the encoder and decoder to initialize their downstream model. However, their method requires a separate diagnostic procedure, since it was trained on a private dataset consisting only of CT examinations from patients with COVID-19.

\subsection{Organ \& Tumour Segmentation}

Multiple studies report results for pancreatic tumour segmentation on a $4$-fold cross validation on the public {\tt NIH Pancreas-CT} dataset~\cite{pancreasct}. Custom pretext tasks for this problem include reconstruction after shuffling CT slices~\cite{zheng_improving_2020}, reconstruction of scrambled subvolumes~\cite{tao_revisiting_2020}, and contrastive learning using inter- and intra-case pairwise relationships~\cite{yang_voxsep_2022}. Table~\ref{tab:nih-pancreas-comparison} compares results reported by these studies. \cite{taleb_3d_2020} and \cite{zhang_sar_2021} report improved segmentation of pancreatic tumours in the public {\tt MSD} dataset~\cite{simpson2019MSD}, compared to training from scratch. 

\begin{table}[h!]
    \centering
    \begin{tabular}{cccc}
         \toprule
         \multicolumn{2}{c}{{\sc Method}} &  \multicolumn{2}{c}{{\sc Initialization}} \\
         \cmidrule(lr){1-2} \cmidrule(lr){3-4}
         \small
         First Author [ref] & Identifier &  Random & SSL \\
         \midrule
         Zheng \cite{zheng_improving_2020} & Slice Shuffle & $0.8569$ & $0.8621$ \\ 
         Tao \cite{tao_revisiting_2020} & Rubik's cube++ & $0.8209$ & $0.8408$ \\
         Yang \cite{yang_voxsep_2022} & VoxSeP (3D) & $0.8353$ & $0.8571$ \\ 
         \bottomrule
    \end{tabular}
    \caption{A comparison of SSL pretraining studies for segmentation in {NIH Pancreas-CT}~\cite{pancreasct}. The mean Dice score on the standard $4$-fold cross validation is reported.}
    \label{tab:nih-pancreas-comparison}
\end{table}

The {LiTS2017} dataset is a publicly available benchmark for liver tumour segmentation~\cite{bilic2023LiTS2017}. Multiple studies have utilized it to trial their SSL approaches, including the following aforementioned works: Models Genesis~\cite{zhou_models_2021}, Parts2Whole~\cite{feng_parts2whole_2020}, United~\cite{guo_discriminative_2022}, and self-supervised StyleGAN~\cite{hochberg_self_2022}. Table~\ref{tab:liver-segmentation-comparison} compares the intersection over union (IoU) reported by the first three studies -- \cite{hochberg_self_2022} formulates the {LiTS2017} benchmark as a classification task and observed an improvement in AUC when pretraining with their StyleGAN-based approach. Table~\ref{tab:liver-segmentation-comparison} gives strong evidence in favour of pretraining 3D CNNs for liver tumour segmentation. However, once again the 2D CNNs pretrained using Models Genesis on {\tt LUNA2016} were not superior to fully supervised 2D CNNs initialized with {\tt ImageNet}-pretrained weights.

\begin{table}[h!]
    \centering
    \begin{tabular}{cccc}
         \toprule
         \multicolumn{2}{c}{{\sc Method}} &  \multicolumn{2}{c}{{\sc Initialization}} \\
         \cmidrule(lr){1-2} \cmidrule(lr){3-4}
         \small
         First Author [ref] & Identifier &  Random & SSL \\
         \midrule
         Zhou \cite{zhou_models_2021} & Models Genesis (3D) & $0.7976$ & $0.8510$ \\
         Feng \cite{feng_parts2whole_2020} & Parts2Whole & $0.7782$ & $0.8670$ \\ 
         Guo \cite{guo_discriminative_2022} & United & $0.7782$ & $0.8653$ \\ 
         \bottomrule
    \end{tabular}
    \caption{A comparison of SSL pretraining studies for liver tumour segmentation using 3D CNNs on the {LiTS2017} benchmark. We display the intersection over union (IoU) reported by each study.}
    \label{tab:liver-segmentation-comparison}
\end{table}

The {\tt BTCV} benchmark contains abdominal CT scans with segmentation labels for $13$ abdominal organs~\cite{landman2015BTCV}. Tang \etal~\cite{tang_self-supervised_2022} pretrained vision transformers using a composite loss with reconstructive, contrastive, and rotation classification terms, following random masking and rotation of CT volumes. They observed that the combination of all regularizers was superior to a subset of them or training from scratch. Jiang \etal~\cite{jiang_self-supervised_2022} applied masked image modelling and self distillation to train vision transformers, evaluating on {\tt BTCV}. Table~\ref{tab:btcv-comparison} compares SSL pretraining approaches that evaluate on the expansive {\tt BTCV} benchmark. In all cases, pretraining appears to outperform training from scratch by a slim margin.

\begin{table}[h!]
    \centering
    \begin{tabular}{cccc}
         \toprule
         \multicolumn{2}{c}{{\sc Method}} &  \multicolumn{2}{c}{{\sc Initialization}} \\
         \cmidrule(lr){1-2} \cmidrule(lr){3-4}
         \small
         First Author [ref] & Identifier &  Random & SSL \\
         \midrule
         Yang \cite{yang_voxsep_2022} & VoxSeP & $0.8428$ & $0.8601$ \\
         Tang \cite{tang_self-supervised_2022} & N/A & $0.8343$ & $0.8472$ \\ 
         Jiang \cite{jiang_self-supervised_2022} & SMIT & $0.8500$ & $0.8778$ \\
         \bottomrule
    \end{tabular}
    \caption{A comparison of SSL pretraining studies for the {\tt BTCV} benchmark. We display the average Dice score across the {\tt BTCV} tasks reported by each study.}
    \label{tab:btcv-comparison}
\end{table}

Zheng \etal~\cite{zheng_hierarchical_2021} trialled a different composite loss for a hierarchical pretext task. They pretrained using multiple datasets, formulating classification losses for contrastive learning, task prediction, and group prediction (where a group is a subset of anatomically similar datasets), along with a reconstruction loss. They argued that the use of these regularizers would facilitate the integration of hierarchical knowledge embodied by the relationship of the datasets to one another into the feature extractor. Through ablation studies, they found that all components of the regularizer led to the best performance and that their approach was superior to a standard encoder-decoder architecture.

Lastly, multiple studies have observed improvement for self-supervised pretraining for the task of organ-at-risk segmentation, which plays a vital role in radiotherapy. Pretext tasks included multi-view momentum contrastive learning~\cite{liu_multiview_2021}, predicting inter-slice distance~\cite{yu_segmentation_2022}, and an extension of Models Genesis with patch swapping~\cite{francis_sabos-net_2022}. However, the experimental validity of \cite{liu_multiview_2021} is severely limited because the test set was included in the dataset used for pretraining.

\subsection{Other CT Diagnostic Tasks}

A plethora of studies investigate self-supervised pretraining for a variety of diagnostic tasks on CT, demonstrating its merit. Examples of other tasks explored include kidney tumour classification (with the public {\tt KiTS19} dataset~\cite{heller2019kits19})~\cite{zhao_unsupervised_2021}, liver lesion classification~\cite{dong_case_2021, ma_self-supervision_2021}, renal cell carcinoma grading~\cite{xu_deep_2022}, grading of non-alcoholic fatty liver disease~\cite{jana_liver_2021}, object detection for lesions~\cite{shou_object_2022} and organs~\cite{chen_self-supervised_2019}, coronary vessel segmentation~\cite{kraft_overcoming_2021}, whole heart segmentation (with the public {\tt CT-WHS} dataset~\cite{zhuang2010ctwhs-mri-whs})~\cite{dong_self-supervised_2021}, abdominal muscle segmentation~\cite{mcsweeney_transfer_2022}, and pneumothorax segmentation~\cite{xue_snu-net_2022}.

\section{Applications in Magnetic Resonance Imaging}

As another 3D modality, MR examinations are cumbersome to segment. Unsurprisingly, there exist several studies that have leveraged self-supervised pretraining to derive value from unlabelled MRI data. Here we enumerate and evaluate evidence regarding the effect of pretraining for diagnostic tasks with MRI.

\subsection{Brain MRI Diagnostic Tasks}
\label{subsec:brain-mri}

Brain tumour segmentation is a frequently studied downstream task for which open datasets exist. The {\tt BraTS} challenge~\cite{menze2014BraTS} is a common benchmark for multi-modal MRI segmentation. It contains anatomically aligned T1, contrast T1, T2, and FLAIR brain MR scans, along with ground truth segmentation labels for brain tumours. The {\tt BraTS} dataset has been updated multiple times, and is often referred to in conjunction with the year the challenge was held.  Several reconstructive pretext tasks have been proposed for this problem. Chen \etal~\cite{chen_self-supervised_2019} adopted a reconstructive task, corrupting MRI slices by swapping locations of square patches of pixels. They observed an improvement in nearly all performance metrics when using $25\%$ and $50\%$ of the dataset. However, they did not perform a comparison using all of the available training labels. Kayal \etal~\cite{kayal_region--interest_2020} presented an inpainting pretext task where 3D supervoxels were redacted from the volume. Their 3D CNN significantly outperformed randomly initialized baselines when pretrained using their self-supervised objective, even when all training labels were included. Expanding on the jigsaw pretext task, Taleb \etal~\cite{taleb_multimodal_2021} demonstrated that including multiple MRI modalities in the pretraining phase was an improvement from single-modality pretraining and training from scratch. Since {\tt BraTS} is multi-modal, it is unsurprising that representations from a single-modality pretrained network would trail multi-modality pretraining. They also applied generative methods to produce patches for underrepresented modalities. The patches used to construct the jigsaw puzzles were composed of segments from different modalities. In an effort to improve feature representations for boundary regions (and therefore downstream segmentation quality), Huang \etal~\cite{huang_attentive_2022} adopted a standard cuboid patch masking reconstructive task using a vision transformer, but applied a weighting factor to voxels belonging to regions where the intensity is rapidly changing. They also applied a symmetric position encoding that ensured equivalence of position encoding for corresponding left and right sides of the brain. An ablation study highlighted the merit of both of these improvements, evaluating on {\tt BraTS 2021}. Unfortunately, it is difficult to compare the results highlighted by the aforementioned techniques because they were evaluated on different editions of {\tt BraTS}.

A substantial number of studies have focused on using self-supervised pretraining for the detection of psychiatric diseases. Several studies have utilized the {\tt ADNI}~\cite{mueller2005ADNI} and {\tt OASIS}~\cite{marcus2007OASIS} datasets to develop classifiers that can detect brain MRI volumes of patients with Alzheimer's disease (AD). Mahmood \etal developed 1D CNNs on time courses of resting state fMRI examinations to detect AD, schizophrenia, and autism. They pretrained using a contrastive pretext task where the pairwise relationship consisted of a fragment of and the entirety of a time course, which improved AUC for all three classifiers. Fedorov~\etal~\cite{fedorov_self-supervised_2021, fedorov_self-supervised_2021-1} proposed contrastive pretraining where positive pairs consisted of paired fMRI and T1 MRI frames. The results are mixed, with fully supervised models outperforming pretrained models for T1 volumes and vice versa for fMRI volumes, for the task of AD detection. Leveraging the multiple examples per patient available in {\tt ADNI}, Zhao \etal~\cite{zhao_longitudinal_2021} suggested a pretext task that combines a basic autoencoder with mean squared error with a regularizer intended to enforce directionality in the latent space for representations of volumes from the same patient taken at two points in time. The regularizer maximizes the cosine between the difference between the representations of paired newer and older examples and a constant vector, $\mathbf{\tau}$. The idea is to learn representations such that adding a scalar multiple of $\mathbf{\tau}$ corresponds to an increase in brain age. Decoded MR examples that varied along $\mathbf{\tau}$ indicated morphological differences associated with increased brain age. Lastly, pretrained models performed better than those initialized randomly. Expanding on this approach, these authors proposed a pretext task that clusters examinations with similar brain age, while still enforcing a direction of increasing brain age within neighbourhoods~\cite{ouyang_self-supervised_2022}. Their approach improved test AUC by $0.076$ compared to their previous work. Dufumier \etal~\cite{dufumier_contrastive_2021} report an improvement on AD detection in {\tt ADNI} over full supervised learning, pretraining with Models Genesis~\cite{zhou_models_2021}, and SimCLR~\cite{chen2020simple} when incorporating a weight into the standard contrastive objective corresponding to the difference in a continuous meta-variable, such as patient age. Other pretexts that have demonstrated improved performance in AD detection include contrastive learning with positive pairs composed from different orthogonal slice views and variable-length volumes~\cite{cao_multiview_2022}, and positive pairs composed by pasting anatomically bounded components of one image onto another~\cite{seyfioglu_brain-aware_2022}. Moving the focus away from pathology, Osin \etal~\cite{osin_learning_2020} were able to train a linear classifier using representations provided by a feature extractor pretrained to predict next-frame amygdala activity on fMRI. The classifier performed better than a CNN baseline at predicting demographic traits (e.g., age) and psychiatric traits according to clinical questionnaires (e.g., trait anxiety).

SSL has also proved useful for automatic white matter segmentation. In 2020, Lu \etal~\cite{lu_white_2020} devised a pretext task for white matter segmentation on diffusion MRI (dMRI) images from the openly available {\tt Human Connectome Project}~\cite{van2013HCP} that involved predicting density maps of white matter fiber streamlines. The labels for this pretext task were generated by applying a previously proposed tractography algorithm and producing a density map by aggregating the number of streamlines intersecting each voxel. After fine-tuning, the Dice score of the pretrained model was $0.137$ greater than that of randomly initialized model. The following year, Lu \etal~\cite{lu_volumetric_2021} extended this work by introducing a second pretext task that involved segmenting white matter based on labels computed using a registration-based algorithm available in a separate software package. They optimized the feature extractor on the first and then second pretext task (i.e., sequentially). Models pretrained using either or both of the pretext tasks outperformed the baseline. Interestingly, the authors did not compare sequential pretraining with simultaneous optimization of both objectives using separate decoder heads. Huang \etal~\cite{huang_attentive_2022} also applied their method (see previous paragraph) to the downstream task of white matter segmentation on the publicly available {\tt WMH} dataset~\cite{kuijf2019WMH}, but did not compare with a fully supervised baseline.

Studies have witnessed performance gains for other brain MRI tasks, such as brain anatomy segmentation~\cite{chang_boundary-enhanced_2022, zoetmulder_domain-_2022, tran_ss-3dcapsnet_2022}, multiple sclerosis lesion segmentation~\cite{cao_multiview_2022, zoetmulder_domain-_2022}, and stroke lesion segmentation~\cite{zoetmulder_domain-_2022}. For instance, in an effort to improve brain anatomy segmentation, Chang \etal pretrained to solve two pretext tasks: (1) predicting the location of the vocal in the nearest supervoxel and (2) predicting the deformation field between the current volume and an atlas. Similar to \cite{huang_attentive_2022}, the first term promotes saliency in rapidly changing regions close to boundaries. The second term requires the encoder to produce features that highlight boundaries of larger structures, which are required for a registration task. Zoetmulder \etal~\cite{zoetmulder_domain-_2022} assessed the utility of supervised and self-supervised pretraining (with an auto-encoding pretext task) for multiple sclerosis lesion segementation, stroke lesion segmentation, and brain anatomy segmentation. They found that pretraining using MRI data resulted in better performance on downstream tasks than with natural images. They did not find that self-supervised pretraining was superior to supervised pretraining for all downstream tasks, but they employed a pretraining dataset that included classification and segmentation labels. While this is an important finding, the major utility of self-supervised learning is to leverage \textit{unlabelled} data when labels are not available.

\subsection{Prostate MRI Diagnostic Tasks}

The prostate segmentation task in {\tt MSD}~\cite{simpson2019MSD} is a benchmark for prostate semantic segmentation, where the task is to segment the peripheral zone and central gland of the prostate. Chaitanya \etal~\cite{chaitanya_contrastive_2020} applied a two-stage pretraining strategy, where an encoder is trained using standard contrastive learning in the first phase, and some decoder blocks are trained during the second phase to minimize a local contrastive loss that encourages dissimilarity among distinct patches in the same image. When fine-tuning, they appended the remainder of the decoder blocks, achieving greater Dice scores than randomly initializing the full model. Taleb \etal~\cite{taleb_multimodal_2021} (see Section~\ref{subsec:brain-mri}) also evaluated their approach using this dataset, but used a different test split.

The {\tt ProstateX} benchmark dataset~\cite{armato2018ProstateX} contains segmentation maps for cancerous lesions of the prostate. Fernandez-Quilez \etal~\cite{fernandez-quilez_contrasting_2022} observed that pretraining with SimCLR~\cite{chen2020simple} improved downstream segmentation performance, compared random and {\tt ImageNet}-pretrained initialization. They tailored the original stochastic transformation distribution such that it entailed plausible prostate MRI slices. Wang \etal~\cite{wang_self-supervised_2022} engineered a more complex pretext task for the same downstream task, which involved optimizing a contrastive learning objective where images from the same patient comprise a positive pair, and an augmentation classification objective. The two methods cannot be compared because they employed different evaluation protocols. Bolous \etal~\cite{bolous_clinically_2021} also observed an improvement for the same downstream task with a private dataset when pretraining using a reconstructive pretext task.

\subsection{Cardiac MR Segmentation}

Segmentation of cardiac structures with machine learning is an extensively studied topic. Bai \etal~\cite{bai_self-supervised_2019} exploited the orientation of short-axis and long-axis planes as given in DICOM files to create a pretext task consisting of segmentation of fixed-size boxes placed at specific points along lines corresponding to bisection with other axes. The relative locations of the boxes is constant with respect to the major cardiac structures. Pretraining improved downstream segmentation of the left ventricle, right ventricle, and myocardium. Notably, optimizing the pretext and downstream objective simultaneously (in a semi-supervised fashion) yielded the greatest test Dice score. This study constitutes another successful example of leveraging domain knowledge available in unlabelled data. Ouyang \etal~\cite{ouyang_self-supervision_2020} demonstrated that self-supervised learning can replace standard training with labelled images for few-shot segmentation. They constructed superpixels from images and used randomly transformed copies of the original image for both support and query. Remarkably, self-supervised training resulted in better downstream segmentation of the left ventricle, right ventricle, and the myocardium on the {\tt Card-MRI} dataset~\cite{zhuang2018card-mri}. Other studies have integrated SSL into federated learning regimes~\cite{wu_federated_2021} and meta-learning~\cite{kiyasseh_segmentation_2021}, citing improvement in performance for cardiac structure segmentation.

SSL has also proven useful for disease classification on cardiac MRI. Zhong \etal~\cite{zhong_self-supervised_2021} found that, when corrupting cine cardiac MR volumes with random pixel shuffling, patch obfuscation, and entire frame dropout, reconstructive pretraining improved downstream classification of preserved versus reduced ejection fraction subtypes of heart failure. An ablation study demonstrated pretraining using each corruption, in isolation, also improved performance. 

\subsection{Grading Intervertebral Disc Degeneration}

SSL pretraining has been applied successfully to a constellation of other tasks involving MRI data. One of the earliest studies employing SSL for MRI was conducted in 2017 by Jamaludin \etal~\cite{jamaludin_self-supervised_2017}, in which they pretrained a CNN on a spinal MRI dataset for the downstream task of grading disc degeneration disease according to the Pfirrmann system~\cite{pfirrmann2001magnetic}. They pretrained to simultaneously solve two pretext tasks: (1) contrastive learning where positive pairs were longitudinal samples from the same patient and (2) classification of vetebral body level. The pretrained models consistently outperformed models trained from scratch, for varying levels of training label availability. Solving the same downstream task on a different private dataset, Kuang \etal~\cite{kuang_spinegem_2021} adopted a reconstructive pretext task, where inputs were distorted by applying different stochastic transformations to image regions corresponding to vertebral bodies, intervertebral discs, and the background. They used a previously described unsupervised segmentation algorithm to compute masks corresponding to these classes, avoiding the need for labels.

\subsection{Other MR Diagnostic Tasks}

Studies have observed improved downstream performance when conducting self-supervised pretraining for other tasks with MRI data, including intracranial hemorrhage detection~\cite{nguyen_self-supervised_2020}, anterior cruciate ligament tear detection~\cite{atito_sb-ssl_2022}, spinal tumour subtype classification~\cite{jiao_self-supervised_2022}, and abdominal organ segmentation~\cite{nguyen_semi-supervised_2021,jiang_self-supervised_2022}.

\section{Applications in Ultrasound Imaging}

Evidence exists in support of pretraining machine learning models for diagnostic tasks with ultrasound (US) examinations. However, as outlined in Section~\ref{sec:search-methodology}, considerably fewer publications have explored self-supervised pretraining for US than for the preceding three types of radiological imaging. Although US examinations are typically represented as 3D tensors (4D when motion is displayed with colour), they are fundamentally different from CT and MRI in that the third dimension is temporal as opposed to spatial. However, like CT and MRI, there are occasions where a single image is sufficient to perform a particular diagnostic task.

\subsection{US Breast Malignancy Detection}

Nguyen \etal~\cite{nguyen_semi-supervised_2021} explored the efficacy of BYOL~\cite{grill2020bootstrap} for the classification of breast US images from the public {\tt BreastMNIST} dataset~\cite{medmnistv2} as either normal, containing benign tumours, or containing malignant tumours. Although the paper is rife with terminological errors, it provides a benchmark for a nonconstrastive method on a public dataset. They found that pretraining with BYOL resulted in worse test performance than randomly initialized or {\tt ImageNet}-pretrained weights. Perek \etal~\cite{perek_self_2021} arrived at a similar conclusion when trialling MoCo with a private dataset~\cite{he2020momentum}. Proposing a video-specific pretext task instead, Lin \etal~\cite{lin_masked_2022} pretrained an encoder-decoder architecture to restore a US video after randomly masking out entire frames and patches in the remaining frames. Upon performing semi-supervised fine-tuning for benign versus malignant lesion classification on a private dataset, masked video pretraining yielded $1\%$ greater accuracy compared to random initialization. Focusing instead on breast lesion semantic segmentation, Mishra \etal~\cite{mishra_cr-ssl_2022} pretrained an encoder-decoder to perform a deterministic edge detection or segmentation task that does not require machine learning. They performed experiments using two publicly available datasets (BUSI~\cite{al2020busi} \& UDIAT~\cite{yap2017udiat}) and observed that SSL improved performance, with the gap increasing with less labelled training data availability. However, it is unclear which pretext task they selected for their downstream experiments.

\subsection{Echocardiography Tasks}

SSL has been cited as useful for a variety of echocardiography interpretation tasks. Anand \etal~\cite{anand_benchmarking_2022} sought to establish the performance of ubiquitous contemporary joint-embedding SSL methods for the task of view classification (e.g., SimCLR~\cite{chen2020simple}, MoCoV2~\cite{chen2020improved}, BYOL~\cite{grill2020bootstrap}, DINO~\cite{caron2021dino}). Not only did they find that pretraining outperformed random and {\tt ImageNet}-pretrained initialization, but they demonstrated that pretraining with more unlabelled data widened the performance gap. SimCLR and BYOL pretraining have been investigated for the task of left ventricle segmentation. Saeed \etal~\cite{saeed_contrastive_2022} observed that SimCLR pretraining generally resulted in the best Dice score, but the difference was small across label availability fractions. Surprisingly, BYOL pretraining generally resulted in worse performance than full supervision. The results appeared to be consistent across two public datasets: {\tt EchoNet-Dynamic} dataset~\cite{ouyang2020echonet-dynamic} and CAMUS~\cite{leclerc2019camus}. To reduce redundancy of pretraining examples, they chose to use one randomly selected frame per clip during pretraining, despite using two labelled frames per clip for the downsteam task (one each for end-systole and end-diastole); it is possible that using more frames during pretraining may have improved performance. Dezaki \etal~\cite{dezaki_echo-syncnet_2021} devised a multifaceted pretext task customized for echocardiograms that consists of (1) reordering shuffled triplets of contiguous frames, (2) minimizing embeddings for contiuous frames and maximizing embeddings for temporally distance frames, and (3) minimizing the differences between embeddings of frames from multiple views corresponding to the same point in the cardiac cycle. Although fully supervised learning matched self-supervised pretraining when using all labels, SSL greatly improved performance when less labels were available. They observed similar results when evaluating on {\tt EchoNet-Dynamic}. 

\subsection{Assessment of Thyroid Nodules on US}

US is often employed to assess thyroid nodules for possible malignancy. Zhao \& Yang~\cite{zhao_unsupervised_2021} pretrained a classifier to distinguish between benign and malignant nodules, using the public {\tt TN-SCUI2020} dataset. They integrated prior medical knowledge into their contrastive pretext task, which sought to minimize the differences between embeddings of handcrafted radiomics features and the original US image. Their method outperformed random initialization and pretraining with generic pretext tasks. Xiang \etal~\cite{xiang_self-supervised_2022} also devised a custom pretext task for this problem, characterized by thyroid US modality classification. In addition to B-mode US, their downstream model received corresponding images from three US modalities, noting superior performance on their private dataset when pretraining as opposed to random or {\tt ImageNet}-pretrained initialization. Guo \etal~\cite{guo_global_2021} focused on the related downstream task of grading nodules according to the widely adopted TI-RADS~\cite{tessler2017tirads} system.

\subsection{Obstetric US Tasks}

Jiao \etal~\cite{jiao_self-supervised_2020} described a custom US-specific pretext task consisting of predicting the order of $4$ shuffled frames and predicting the continuous parameters of random affine transformations applied to the frames, which resulted in an improvement over training from scratch for the task of fetal plane detection. Chen \etal~\cite{chen_self-supervised_2019} (described in Section~\ref{subsec:brain-mri}) observed a similar result for the same task. Focusing instead on segmenting the utero-placental interface, Qi \etal~\cite{qi_knowledge-guided_2020} pretrained a feature extractor for a customized jigsaw pretext task in which the permuted patches were sampled from image regions intersected the labelled region of interest. Results indicated marginal improvement with pretraining, but the custom pretext task did not outperform Jigsaw~\cite{noroozi2016unsupervised} for the majority of feature extractors studied. Of note is the fact that their pretext task cannot be considered SSL because, by definition, the pretext task is solved in the absence of labels.

\subsection{Other US Diagnostic Tasks}

Liu \etal~\cite{liu_tn-usma_2021} pretrained an encoder-decoder model for the downstream task of classifying gastrointestinal stromal tumours from endoscopic US images, observing greater average performance than full supervision with {\tt ImageNet}-pretrained initialization (albeit with greatly overlapping confidence intervals). Interestingly, they leveraged thyroid and breast US datasets for pretraining. Zhou \etal~\cite{zhou_rating_2022} found random permutation prediction to be a helpful pretext task for rheumatoid arthritis grading on US; however, the approach required manual region of interest labelling. Lastly, Basu \etal~\cite{basu_unsupervised_2022} proposed an US-specific contrastive pretext task that considered temporally separated frames from the same video as negative pairs, in addition to inter-video pairs. Positive pairs were frames separated temporally by no more than a predefined constant number of time steps. Further, they imposed a curriculum by gradually decreasing the minimum temporal distance constituting an intra-video negative pair. Intra-video negative pairs are important to consider because the anatomical context of an US video may differ dramatically throughout its duration. However, the authors did not address how negative pair sampling would be considered for cases where the probe is kept stationary throughout the video. The authors evaluated their approach on a private dataset for gallbladder malignancy detection and on the public {\tt POCOVID-Net}~\cite{born2020pocovid} lung US dataset for COVID-19 classification, citing performance superior to {\tt ImageNet}-pretrained initialization, SimCLR~\cite{chen2020simple}, and MoCoV2~\cite{chen2020improved}.

\section{Assessment \& Future Directions}

\subsection{Evidence for SSL Pretraining}

\subsubsection{Comparison to Random Initialization}

The previous sections of this work illustrate the usefulness of self-supervised pretraining in deep learning for diagnostic tasks with radiological images. For each of the four major modalities investigated, there are multiple studies that report an improvement in downstream performance metrics when initializing feature extractors with SSL-pretrained weights, generally compared to random weight initialization in the fully supervised setting.

In most cases, studies demonstrated that pretraining was useful either as a first step using all labelled data, or that pretraining was particularly helpful in low label availability settings. When labels are completely available for a downstream task, there is a wide variation in the change in performance on test data. Some studies report marginal to no improvement~\cite{liu_tn-usma_2021, qi_knowledge-guided_2020, haghighi_dira_2022}, while others report significant gains~\cite{lu_white_2020, kuang_spinegem_2021, ouyang_self-supervised_2022, atito_sb-ssl_2022}. Naturally, there are myriad reasons for such variability, including dissimilar pretext tasks, evaluation protocol differences, modality-specific noise, dataset volume and diversity, and downstream task difficulty.

The results of this review overwhelmingly suggest that pretraining with self-supervised learning is likely to result in improved performance on downstream supervised learning tasks, compared to randomly initialized supervised learners. Practitioners should consider trialling pretrained feature extractors during model development.

\subsubsection{The Prowess of {\tt ImageNet}-pretrained Weights}

The vast majority of the methods explored in this review compared their pretrained models to the fully supervised setting where weights are randomly initialized. Many also compared the results of their custom SSL method to previously proposed SSL methods that are not geared toward any specific imaging distribution. However, a fraction of studies compared their pretrained feature extractors to the ubiquitously employed {\tt ImageNet}-pretrained weights. It is reasonable to compare against {\tt ImageNet}-pretrained weights because several medical computer vision models are initialized with {\tt ImageNet}-pretrained weights~\cite{azizi_big_2021}. Indeed, many studies reported that {\tt ImageNet}-pretrained weights fared better than random initialization, making them a stronger baseline against which to compare. Crucially, multiple studies reported cases where 2D CNNs or vision transformers did not appreciably outperform {\tt ImageNet}-pretrained initialization~\cite{zhou_models_2021, haghighi_learning_2020, feng_parts2whole_2020, nguyen_semi-supervised_2021, pang_popar_2022, haghighi_dira_2022}. We therefore advise authors of future SSL studies to compare their approaches to fully supervised baselines with random initialization \textit{and} {\tt ImageNet}-pretrained initialization where applicable.

A frequently absent experimental setting is the assessment of the effect of initializing feature extractors with {\tt ImageNet}-pretrained weights \textit{prior} to self-supervised pretraining. The small set of studies that performed this comparison observed that the best performance in downstream radiological imaging interpretation tasks was achieved by setting the initial weights of the feature extractors to {\tt ImageNet}-pretrained weights~\cite{azizi_big_2021, ma_benchmarking_2022}. Future studies should include this experiment in their evaluation protocol.

Of course, it is necessary to acknowledge that publicly available {\tt ImageNet}-pretrained weights do not exist for all feature extractor architectures (e.g., 3D CNNs), and that fully supervised pretraining can be prohibitively expensive.

\subsubsection{Utility of SSL in Low-Label Settings}

Aside from direct comparisons to fully supervised counterparts using the same dataset, several studies have established the benefit of self-supervised pretraining in scenarios where labels are not provided for all available examples. Typically, such claims are established by comparing performance of fully and self-supervised models at different fractions of label availability, limiting the amount of data available for supervised fine-tuning on a downstream task~\cite{chen_self-supervised_2019,taleb_3d_2020, azizi_big_2021, kayal_region--interest_2020, dezaki_echo-syncnet_2021, zheng_hierarchical_2021}. Some studies reported changes in downstream performance when larger unlabelled datasets that dwarfed the available labelled examples were leveraged for pretraining~\cite{zhang_sar_2021,anand_benchmarking_2022, ma_benchmarking_2022}. Some studies even demonstrated that pretraining with unlabelled data geared for a different downstream task but that was collected using the same modality can improve downstream performance~\cite{dufumier_contrastive_2021, liu_tn-usma_2021, lu_self-supervised_2022, huang_attentive_2022}. In the extreme scenario of few-shot learning, self-supervised objectives may be employed during training~\cite{ouyang_self-supervision_2020}. The considerable amount of evidence outlined in this review suggests that practitioners should leverage unlabelled data when available and pretrain feature extractors using SSL.

\subsubsection{Relative Dearth of Ultrasound Research}

As depicted in Figure~\ref{fig:papers-by-modality}, the number of papers eligible for inclusion in this review concerning US data is less than half of the number included for X-ray, CT, or MRI. Hence, there exists a need for (1) more investigations that quantify the impact of preexisting SSL pretraining tasks for US tasks and (2) studies that modify preexisting or propose novel SSL methods that are suited to the US modality. US presents additional challenges for machine learning systems compared to the other modalities, such as increased noise, the temporal dimension, acquisition-related differences in probe movement and orientation, motion artefacts, and geometrical differences across probe types and manufacturers. As a result, further work is warranted in determining the types and aspects of pretext tasks suitable for US.

\subsection{Theoretical Support for Empirically Validated Methods}

The majority of the studies presented in this review provide SSL methods that are presented as task-specific, instead of applying preexisting methods to ne. Some deviate wildly from previous work~\cite{chang_boundary-enhanced_2022, huang_self-supervised_2022, zoetmulder_domain-_2022, seyfioglu_brain-aware_2022}, and others are incremental changes to previously explored pretext tasks~\cite{jana_liver_2021, mcsweeney_transfer_2022}. The pretext tasks put forth in such studies are often fashioned with clinical and/or background knowledge about the downstream task, but are mostly justified by intuition. The arguments for further use of the proposed methods typically consists entirely of empirical validation. Multiple such studies boast superior performance of their methods boast empirical results but do not establish statistically significant improvements~\cite{zhu_rubiks_2020,liu_tn-usma_2021, zheng_msvrl_2022}.

As discussed in Section~\ref{subsec:theoretical-justification}, some SSL methods have received theoretical justification in terms of performance on downstream task. Many studies discussed in this survey employ such justified methods, such as SimCLR~\cite{chen2020simple} and Barlow Twins~\cite{zbontar2021barlow}. These methods are guaranteed to improve performance on downstream tasks as long as the labels for positive pairs would be the same in the downstream task. For example, Fernandez-Quilez \etal~\cite{fernandez-quilez_contrasting_2022} employed SimCLR with a modified transformation distribution that captured differences between positive pairs that would not constitute a change of label. Azizi \etal~\cite{azizi_big_2021} also employed SimCLR, but expanded the pairwise relationship to include multiple acquired views of the same pathology. Applying custom data augmentation transformations that do not change the label distribution in the downstream task or defining the pairwise relationship based on preexisting clinical knowledge are viable strategies for the successful application of theoretically justified joint-embedding SSL methods. Such clinical knowledge may come ``for free" in that it does not require further labelling --- practitioners could consider sources such as multiview examinations, multimodal studies, accompanying radiology reports, and DICOM tags. Future methods should strive to apply theoretically justified approaches to SSL pretraining where possible; otherwise, statistical significance testing should be conducted when claims are made regarding the superiority of novel methods.

\subsection{Comparable and Reproducible Benchmarks}

A longstanding problem in machine learning for medical imaging is the lack of public datasets, which thwarts replicability of results. A considerable number of studies in this review presenting novel SSL methods for radiological imaging tasks conducted their evaluations on private datasets only. As a result, many of the results presented are not directly commparable. This review was only able to directly compare studies for a limited set of downstream tasks where authors reported performance on public datasets. Authors suggesting novel SSL methods are encouraged to evaluate their methods using public datasets, or to include results on public datasets in addition to their private datasets (e.g., \cite{dezaki_echo-syncnet_2021,wang_self-supervised_2022}). When evaluating on public datasets, researchers should use train/test splits that are identical to preceding studies. Furthermore, authors should endeavour to utilize identical pretraining and training sets when evaluating their approach on standard public datasets. To promote usage of public benchmarks in future studies, Tables~\ref{tab:public_datasets_xr},~\ref{tab:public_datasets_ct},~\ref{tab:public_datasets_mr},~and~\ref{tab:public_datasets_us} detail all public datasets referenced in this review, providing URLs for access.

\subsection{The Impact of Pretraining on Generalizability}

Machine learning models trained for tasks involving radiological images are utterly susceptible to performance drops under distributional shift~\cite{varoquaux2022machine}. Biases can be introduced by the distribution of confounding or mediating variables in the training set, such as labelling discrepancies, patient demographics, acquisition technique, and device manufacturer. External validation is therefore a pivotal pre-deployment step. Some studies in this review reported improvement in performance on external test sets when self-supervised pretraining was conducted~\cite{azizi_big_2021, tiu_expert-level_2022, zhou_rating_2022}, but further work is required to confidently characterize this phenomenon.

\section{Conclusions}

This work reviewed a range of recent studies across modalities, datasets, and methods that explored the impact of self-supervised pretraining for the automation of diagnostic tasks in radiological imaging. The consensus observed in the majority of the publications included in this survey suggest that SSL pretraining using unlabelled datasets generally improves the performance of supervised deep learning models for downstream tasks in radiography, computed tomography, magnetic resonance imaging, and ultrasound. The findings substantiate the utility of unlabelled data in radiological imaging, thereby reducing the prohibitive expense of expert labelling. Practitioners should therefore consider self-supervised pretraining when unlabelled data is abundant. Future work in SSL for radiological imaging should focus on developing and/or applying theoretically justified methods that capitalize on clinical knowledge, further exploring SSL for problems in ultrasound, and ascertain the effect of SSL on generalizability.

\section{List of Abbreviations}

\printglossary[type=\acronymtype,title=,nogroupskip=true]


\begin{appendices}

\section{Database Queries}
\label{apx:database-queries}

Four databases were queried in November 2022 to gather the set of candidate studies for review, including Scopus\footnote{\url{https://www.scopus.com}}, IEEE Xplore Digital Library\footnote{\url{https://ieeexplore.ieee.org}}, ACM Digital Library\footnote{\url{https://dl.acm.org}}, and PubMed\footnote{\url{https://pubmed.ncbi.nlm.nih.gov}}. Below are the queries, along with the number of results returned by each database. 

\vspace{0.5em}
\noindent \textbf{Scopus} ($856$ results)
\vspace{0.5em}

\begin{flushleft}
{\small \tt 
         ( \\
         \hspace*{0.2cm} TITLE-ABS-KEY ( ( "self-supervis*" OR \\
         \hspace*{3.8cm} "\{contrastive learn*\}" ) ) \\ 
         \hspace*{0.2cm} AND \\ 
         \hspace*{0.2cm} TITLE-ABS-KEY ( ( "\{*medical imag*\}" OR \\
         \hspace*{3.6cm} "X-ray*"  OR \\
         \hspace*{3.6cm} ultrasound*  OR \\
         \hspace*{3.6cm} "\{computed tomography\}"  OR \\
         \hspace*{3.6cm} "CT"  OR \\
         \hspace*{3.6cm} "\{magnetic resonance imaging\}"  OR \\
         \hspace*{3.6cm} mri ) ) \\ 
         )
}
\end{flushleft}

\vspace{0.5em}
\noindent \textbf{IEEE Xplore Library} ($328$ results)
\vspace{0.5em}

\begin{flushleft}
{\small \tt 
         ((("All Metadata":"self-supervis*" OR \\
         \hspace*{0.4cm} "All Metadata":"contrastive learn*") \\ 
         \hspace*{0.3cm} AND \\ 
         \hspace*{0.3cm} ("All Metadata":"*medical imag*" OR \\
         \hspace*{0.5cm} "All Metadata":"X-ray" OR \\
         \hspace*{0.5cm} "All Metadata":"computed tomography" OR \\
         \hspace*{0.5cm} "All Metadata":CT OR \\
         \hspace*{0.5cm} "All Metadata":"ultrasound" OR \\
         \hspace*{0.5cm} "All Metadata":"magnetic resonance imaging" OR \\
         \hspace*{0.5cm} "All Metadata":MRI)) \\ 
         \hspace*{0.2cm} OR \\
         \hspace*{0.3cm} ("Index Terms":"self-supervis*" \\
         \hspace*{0.4cm} AND \\
         \hspace*{0.4cm} ("Index Terms":"*medical imag*" OR \\
         \hspace*{0.5cm} "Index Terms":"X-ray*" OR \\
         \hspace*{0.5cm} "Index Terms":"computed tomography" OR \\
         \hspace*{0.5cm} "Index Terms":CT OR \\
         \hspace*{0.5cm} "Index Terms":"ultrasound" OR \\
         \hspace*{0.5cm} "Index Terms":"magnetic resonance imaging" OR \\
         \hspace*{0.5cm} "Index Terms":MRI)) \\
         )
}
\end{flushleft}

\vspace{0.5em}
\noindent \textbf{ACM Digital Library} ($8$ results)
\vspace{0.5em}

\begin{flushleft}
{\small \tt 
(\\
\hspace*{0.2cm} Title:("medical image" OR "medical images" OR "medical imaging" \\ 
\hspace*{1.6cm} OR "biomedical image" OR "biomedical images" OR \\
\hspace*{1.6cm} "biomedical imaging" OR x\-ray OR "computed tomography" OR \\
\hspace*{1.6cm} ct OR ultrasound OR "magnetic resonance imaging" OR mri) \\ 
\hspace*{0.2cm} OR \\
\hspace*{0.2cm} Abstract:("medical image" OR "medical images" \\
\hspace*{2.2cm} OR "medical imaging" OR "biomedical image" OR \\ 
\hspace*{2.2cm} "biomedical images" OR "biomedical imaging" OR x\-ray OR \\
\hspace*{2.2cm} "computed tomography" OR ct OR ultrasound OR \\
\hspace*{2.2cm} "magnetic resonance imaging" OR mri) \\
\hspace*{0.2cm} OR \\
\hspace*{0.2cm} Keywords:("medical image" OR "medical images" OR \\ 
\hspace*{2.2cm} "medical imaging" OR "biomedical image" OR \\ 
\hspace*{2.2cm} "biomedical images" OR "biomedical imaging" OR x\-ray OR \\
\hspace*{2.2cm} "computed tomography" OR ct OR ultrasound OR \\
\hspace*{2.2cm} "magnetic resonance imaging" OR mri)\\
) \\ 
AND \\
( \\
\hspace*{0.4cm}Title:"self\-supervised" OR Abstract:"self\-supervised" OR \\
\hspace*{0.2cm} Keywords:"self\-supervised" OR Title:"self\-supervision" OR \\
\hspace*{0.2cm} Abstract:"self\-supervision" OR Keyword:"self\-supervision" \\
)
}
\end{flushleft}

\vspace{0.5em}
\noindent \textbf{PubMed} ($42$ results)
\vspace{0.5em}

\begin{flushleft}
{\small \tt 
( \\
\hspace*{0.2cm} self-supervis*[tiab] OR "self supervis*"[tiab] OR \\
\hspace*{0.2cm} "contrastive learn*"[tiab] \\
) \\
AND \\
( \\
\hspace*{0.2cm} Machine Learning[Mesh:NoExp] OR deep learning[mesh:noex] \\
) \\ 
AND \\
( \\
\hspace*{0.2cm} "medical imag*"[tiab] OR diagnostic imaging[mesh] OR mri[tiab] OR \\ 
\hspace*{0.2cm} ct[tiab] OR "computed tomography"[tiab] OR "x-ray"[tiab] OR \\
\hspace*{0.2cm} ultrasound[tiab] \\
)
}
\end{flushleft}

\section{Public Datasets}
\label{apx:public-datasets}

Several studies reviewed in this work utilized public datasets. To promote replicability via the usage of public benchmarks, we provide brief descriptions and links for these datasets. Tables~\ref{tab:public_datasets_xr},~\ref{tab:public_datasets_us},~\ref{tab:public_datasets_ct}, and~\ref{tab:public_datasets_mr} list the public datasets for X-ray, US, CT, and MRI respectively.

\begin{table}[h!]
    \centering
    \begin{tabular}{cL{8.5}}
        \toprule
         Name [Citation] & Description \\
         \midrule \addlinespace[0.5em]
         {\tt \href{https://stanfordmlgroup.github.io/competitions/chexpert/}{CheXpert}}~\cite{irvin2019chexpert} & A fully manually annotated $14$-class dataset of chest X-rays. \\ \addlinespace[0.5em]
         {\tt \href{https://nihcc.app.box.com/v/ChestXray-NIHCC}{ChestX-ray14}}~\cite{wang2017chestxray14}  & A $14$-class dataset of chest X-rays with labels extracted from radiology reports. \\ \addlinespace[0.5em]
         {\tt \href{https://www.rsna.org/education/ai-resources-and-training/ai-image-challenge/rsna-pneumonia-detection-challenge-2018}{RSNA Pneumonia}}~\cite{rsna-pneumonia-detection-challenge} & Chest X-rays with bounding box labels for bacterial and viral pneumonias \\ \addlinespace[0.5em]
         {\tt \href{https://medmnist.com}{ChestMNIST}}~\cite{medmnistv2}  & Identical to {\tt \href{https://nihcc.app.box.com/v/ChestXray-NIHCC}{ChestX-ray14}}. Part of {\tt \href{https://medmnist.com}{MedMNIST}}. \\ \addlinespace[0.5em]
         {\tt \href{https://medmnist.com}{PneumoniaMNIST}}~\cite{medmnistv2}  & Paediatric chest X-rays labelled for the presence or absence of pneumonia. Part of {\tt \href{https://medmnist.com}{MedMNIST}}.\\ \addlinespace[0.5em]
         {\tt \href{https://www.kaggle.com/datasets/andyczhao/covidx-cxr2}{COVIDx CXR-2}}~\cite{zhao2021covidx}  & chest X-rays labelled for the presence or absence of COVID-19. \\ \addlinespace[0.5em]
         {\tt \href{https://physionet.org/content/mimic-cxr-jpg/2.0.0}{MIMIC-CXR}}~\cite{johnson2019mimic}  & Chest X-rays, metadata, and free text reports. Same label categories as {\tt CheXpert}. Some labels were manually determined, and others were automatically assigned using the reports. \\ \addlinespace[0.5em]
         \bottomrule
    \end{tabular}
    \caption{The public X-ray datasets referenced in this review, including links to request or download the data.}
    \label{tab:public_datasets_xr}
\end{table}

\begin{table}[h!]
    \centering
    \begin{tabular}{cL{8.5}}
        \toprule
         Name [Citation] & Description \\
         \midrule \addlinespace[0.5em]
        {\tt \href{https://wiki.cancerimagingarchive.net/pages/viewpage.action?pageId=1966254}{LIDC-IDRI}}~\cite{armato2011LIDC-IDRI}  & Chest CT exams labelled for lung nodule classification and segmentation. \\ \addlinespace[0.5em]
        {\tt \href{https://luna16.grand-challenge.org}{LUNA2016}}~\cite{setio2017luna2016} & Chest CT exams labelled for the presence of lung nodules. \\ \addlinespace[0.5em]
        {\tt \href{https://www.rsna.org/education/ai-resources-and-training/ai-image-challenge/rsna-pe-detection-challenge-2020}{RSNA-PE}}~\cite{colak2021RSNA-PE} &  Chest CT exams annotated with instances of pulmonary emboli. \\ \addlinespace[0.5em]
        {\tt \href{https://github.com/UCSD-AI4H/COVID-CT}{COVID-CT}}~\cite{zhao2020covid-ct}  & Chest CT exams labelled for the presence or absence of COVID-19. \\ \addlinespace[0.5em]
        {\tt \href{https://wiki.cancerimagingarchive.net/display/Public/Pancreas-CT}{NIH Pancreas-CT}}~\cite{pancreasct}  & Abdominal contrast-enhanced CT scans with pancreas segmentation labels \\ \addlinespace[0.5em]
        {\tt \href{http://medicaldecathlon.com}{MSD Pancreas}}~\cite{simpson2019MSD} &   Abdominal CT exams with segmentation labels for pancreas parenchyma, cysts, and tumours. Part of the Medical Segmentation Decathlon. \\ \addlinespace[0.5em]
        {\tt \href{https://www.synapse.org/#!Synapse:syn3193805/wiki/217789}{BTCV}}~\cite{landman2015BTCV}  & Abdominal CT exams with segmentation labels for $13$ organs. \\ \addlinespace[0.5em]
        {\tt \href{https://kits19.grand-challenge.org}{KiTS19}}~\cite{} & CT exams labelled for kidney tumour segmentation  \\ \addlinespace[0.5em]
        {\tt \href{http://www.sdspeople.fudan.edu.cn/zhuangxiahai/0/mmwhs}{CT-WHS}}~\cite{zhuang2010ctwhs-mri-whs} &  Axial CT exams with segmentation labels for the ventricles and atria of the heart. \\ \addlinespace[0.5em]
         \bottomrule
    \end{tabular}
    \caption{The public CT datasets referenced in this review, including links to request or download the data.}
    \label{tab:public_datasets_ct}
\end{table}

\begin{table}[h!]
    \centering
    \begin{tabular}{cL{8.5}}
        \toprule
         Name [Citation] & Description \\
         \midrule \addlinespace[0.5em]
        {\tt \href{http://braintumorsegmentation.org}{BraTS}}~\cite{menze2014BraTS} &   MRI exams labelled for brain tumour segmentation and classification. The benchmark has been updated and previous versions are available. \\ \addlinespace[0.5em]
        {\tt \href{https://adni.loni.usc.edu}{ADNI}}~\cite{mueller2005ADNI} & Brain MRI exams with labels for normal controls, mild cognitive impairment, and Alzheimer's disease   \\ \addlinespace[0.5em]
        {\tt \href{https://oasis-brains.org}{OASIS}}~\cite{marcus2007OASIS} &  Brain MRI exams with segmentation labels, patient characteristics, and labels for Alzheimer's disease  \\ \addlinespace[0.5em]
        {\tt \href{https://www.humanconnectome.org}{HCP}}~\cite{van2013HCP} & Unannotated multi-modal MR scans \\ \addlinespace[0.5em]
        {\tt \href{https://dataverse.nl/dataset.xhtml?persistentId=doi:10.34894/AECRSD}{WMH}}~\cite{kuijf2019WMH} & Brain MRI exams with labels for white matter hyperintensities \\ \addlinespace[0.5em]
        {\tt \href{https://wiki.cancerimagingarchive.net/pages/viewpage.action?pageId=23691656}{ProstateX}}~\cite{armato2018ProstateX} & Prostate MRI studies labelled for localization and classification of prostate lesions  \\ \addlinespace[0.5em]
        {\tt \href{https://zmiclab.github.io/zxh/0/mscmrseg19}{Card-MRI}}~\cite{zhuang2018card-mri} & Cardiac MRI exams with labels for ventricular blood volume and myocardium segmentation \\ \addlinespace[0.5em]
         \bottomrule
    \end{tabular}
    \caption{The public MRI datasets referenced in this review, including links to request or download the data.}
    \label{tab:public_datasets_mr}
\end{table}

\begin{table}[h!]
    \centering
    \begin{tabular}{cL{8.5}}
        \toprule
         Name [Citation] & Description \\
         \midrule \addlinespace[0.5em]
        {\tt \href{https://medmnist.com}{BreastMNIST}}~\cite{medmnistv2} &  Breast US images with classification labels for normal, benign lesion, and malignant lesions. \\ \addlinespace[0.5em]
        {\tt \href{https://echonet.github.io/dynamic}{EchoNet-Dynamic}}~\cite{ouyang2020echonet-dynamic} &  Echocardiography videos with end diastolic and end systolic volume labels \\ \addlinespace[0.5em]
        {\tt \href{https://tn-scui2020.grand-challenge.org}{TN-SCUI2020}}~\cite{ouyang2020echonet-dynamic} &  Thyroid US videos with segmentation labels for thyroid nodules \\ \addlinespace[0.5em]
        {\tt \href{https://github.com/jannisborn/covid19_ultrasound}{POCOVID-Net}}~\cite{born2020pocovid} &  Links to lung US videos  labelled for COVID-19, other viral pneumonia, bacterial pneumonia, and healthy lung. \\ \addlinespace[0.5em]
         \bottomrule
    \end{tabular}
    \caption{The public US datasets referenced in this review, including links to request or download the data.}
    \label{tab:public_datasets_us}
\end{table}

\end{appendices}


\clearpage
\bibliographystyle{unsrt}  
\bibliography{references}

\end{document}